\documentclass[]{llncs}
\usepackage{graphicx}
\usepackage{hyperref}
\usepackage{booktabs} 
\usepackage{tabularx}
\usepackage{amsmath}
\usepackage[inline]{enumitem}
\usepackage[pdftex,dvipsnames]{xcolor}  

\usepackage[colorinlistoftodos,prependcaption]{todonotes}
\usepackage{pdflscape}
\usepackage{lscape}
\usepackage{colortbl}
\usepackage{array,multirow}

\usepackage[final]{changes}

\definecolor{subtraj}{HTML}{ffa52f} 
\definecolor{anomaly}{HTML}{6b004f} 
\definecolor{arrival}{HTML}{8c3bff} 
\definecolor{location}{HTML}{018700} 
\definecolor{nextloc}{HTML}{d60000} 
\definecolor{synth}{HTML}{ff7ed1} 
\definecolor{volume}{HTML}{00acc6} 
\definecolor{traj}{HTML}{97fe00} 

\usepackage{fancyhdr}
\fancypagestyle{firstpage}{
    \chead{Manuscript under review at Geoinformatica} 
}

\begin{document}
\title{MobilityDL: A Review of Deep Learning From Trajectory Data}

\titlerunning{Deep Learning From Trajectory Data}
%
\author{Anita Graser\inst{1}\orcidID{0000-0001-5361-2885} \and
Anahid Jalali\inst{1}\orcidID{0000-0002-8889-0735} 
\and
Jasmin Lampert\inst{1}\orcidID{0000-0002-0414-4525}
\and
Axel Weißenfeld\inst{1}\orcidID{0000-0002-7246-2744}
\and
Krzysztof Janowicz\inst{2}
}
\authorrunning{A. Graser et al.}
%
\institute{AIT, Vienna, Austria 
\email{anita.graser@ait.ac.at} \and
University of Vienna, Vienna, Austria}
\maketitle       

\thispagestyle{firstpage}

\begin{abstract}
Trajectory data combines the complexities of time series, spatial data, and (sometimes irrational) movement behavior. 
As data availability and computing power have increased, so has the popularity of deep learning from trajectory data. 
This review paper provides the first comprehensive overview of  deep learning approaches for trajectory data. We have identified eight specific mobility use cases which we analyze with regards to the deep learning models and the training data used.  
Besides a comprehensive quantitative review of the literature since 2018, the main contribution of our work is the data-centric analysis of recent work in this field, placing it along the mobility data continuum which ranges from detailed dense trajectories of individual movers (quasi-continuous tracking data), to sparse trajectories (such as check-in data), and aggregated trajectories (crowd information). 

\end{abstract}

\section{Introduction}

Deep learning (DL) has become a popular approach for developing data-driven prediction, classification, and anomaly detection solutions. Work on deep learning  from trajectory data is spread out over many domains, including but not limited to computer science~\cite{yin_deep_2021,derrow-pinion_eta_2021,feng_deepmove_2018}, geographic information science~\cite{fan_online_2022,gao_traffic_2022,hong_how_2022}, intelligent transportation systems~\cite{singh_leveraging_2022}, maritime sciences~\cite{chen_ship_2020,yang_ship_2022}, and ecology~\cite{lippert_learning_2022}. Consequently, the resulting body of work covers many different use cases and trajectory dataset types.
A data-centric way to categorize trajectory datasets is according to their level of detail~\cite{werner_exploratory_2021} along the mobility data continuum ranging from detailed dense trajectories (quasi-continuous tracking data of individual movement) to sparse trajectories (such as check-in data of individuals), and finally, aggregated trajectories (crowd-level information, typically aggregated to edges/nodes in a mobility graph, to a grid, or to a set of points of interest)\cite{andrienko_conceptual_2011,konzack_trajectory_2018,zheng_trajectory_2015}. 
Even though the level of detail of the underlying trajectory-based training data is an important factor determining the potential capabilities as well as the scale or resolution of derived machine learning models, cursory scanning of paper titles and abstracts is usually insufficient to determine which type of trajectory data was used to train the presented deep learning models. To gain a better understanding of the state of the art, it is instead necessary to review the data and methods sections in detail. While many papers start with dense trajectories, most convert them into sparse trajectories \cite{altan_discovering_2022,chen_ship_2020,fan_online_2022,hong_how_2022,musleh_towards_2022,yang_classification_2018,yang_ship_2022} or even aggregate them to crowd-level \cite{buroni_tutorial_2021,wang_periodic_2022,zhang_curb-gan_2020}. 
Common approaches to turning dense trajectories into sparse trajectories include:
converting them into a sequence of stop locations \cite{altan_discovering_2022,hong_how_2022} or a sequence of traversed regions (grid cells) \cite{fan_online_2022,musleh_towards_2022}, or converting them to trajectory images \cite{chen_ship_2020,yang_classification_2018,yang_ship_2022}.

The recent literature includes review papers for specific mobility use cases, for example, in their review of location encoding methods for GeoAI \cite{mai_review_2022}, the authors stress the analogy between NLP word-to-sentence relations and location-to-trajectory relations. This analogy has led to Word2Vec-inspired approaches encoding location into a location embedding using, for example, Location2Vec \cite{zhu_location2vec_2019}, Place2Vec \cite{yan_itdl_2017}, or POI2Vec \cite{feng_poi2vec_2017}. 
Another recent review \cite{kashyap_traffic_2022} is dedicated to deep learning for traffic flow prediction models, which are primarily trained on aggregated trajectory data.
Finally, Wang et al.~\cite{wang_deep_2022} provide a survey of deep learning for the wider field of spatio-temporal data mining, including trajectories.
However, to the best of our knowledge, there is no review paper that provides a comprehensive overview of the different neural network (NN) architectures used to learn from trajectory data, specifically for mobility use cases.

The goal of this work, therefore, is to provide a first overview of the current state of deep neural networks trained with trajectory data, structured by: 
\begin{enumerate*}
\item Use case category (travel time, crowd flow, and location predictions; location and trajectory classifications; as well as anomaly detection and synthetic data generation),
\item Neural network architecture (including, for example, CNN, RNN, LSTM, and GNN), and
\item Trajectory data granularity (dense, sparse, aggregated) and representation. 
\end{enumerate*}
For practical reasons, the in-depth qualitative assessment part of this paper does not cover all relevant works published in recent years exhaustively. However, we provide at least one paper for each use case and DL combination that we have identified. 
We specifically reviewed publications at recent events, including 
SIGSpatial 2022\footnote{\url{https://sigspatial2022.sigspatial.org/accepted-papers/}}~\cite{fan_online_2022,hong_how_2022,lyu_plug-memory_2022,musleh_towards_2022,tenzer_meta-learning_2022,wang_periodic_2022,xue_leveraging_2022,zhang_factorized_2022},
Sussex-Huawei Locomotion (SHL) Challenge 2021\footnote{\url{http://www.shl-dataset.org/activity-recognition-challenge-2021/}}  at the ACM international joint conference on pervasive and ubiquitous computing (UbiComp)~\cite{wang_locomotion_2021},
Traffic4cast challenge 2021\footnote{\label{Traffic4cast}\url{https://www.iarai.ac.at/traffic4cast/2021-competition/}} at NeurIPS~\cite{lu_learning_2021}, and
Big Movement Data Analytics workshop BMDA 2021\footnote{\url{https://www.datastories.org/bmda21/BMDA21Accepted.html}} at EDBT~\cite{buroni_tutorial_2021,liatsikou_denoising_2021,tritsarolis_online_2021}.

Even though we focus explicitly on deep learning, it is worth noting that deep learning may not always be the best approach~\cite{grinsztajn_why_2022} to address a particular challenge. For example, the SHL Challenge summary~\cite{wang_locomotion_2021} reports that classic machine learning (ML) models outperform deep learning models on their three metrics: F1 score, train time, and test time. In Section~\ref{sec:motivation} we therefore review the main reasons stated by authors to motivate the use of deep learning as well as the baselines and metrics used in evaluations.

This review does not attempt to compare the performance of different deep learning approaches. Even though there are some commonly used open datasets (such as the Porto taxi\footnote{\label{footnote-porto}\url{https://www.kaggle.com/c/pkdd-15-predict-taxi-service-trajectory-i/data}}, the T-Drive taxi\footnote{\label{footnote-tdrive}\url{https://www.microsoft.com/en-us/research/publication/t-drive-trajectory-data-sample/}}, the GeoLife\footnote{\label{footnote-geolife}\url{https://www.microsoft.com/en-us/research/publication/geolife-gps-trajectory-dataset-user-guide/}}, and the Gowalla check-in datasets\footnote{\label{footnote-gowalla}\url{http://snap.stanford.edu/data/loc-Gowalla.html}}) cross-paper comparisons outside of dedicated data challenges \added{(such as the DEBS Grand Challenge 2018~\cite{gulisano_debs_2018}, SHL Challenge 2021~\cite{wang_locomotion_2021}), and Traffic4cast challenge 2021$^{\ref{Traffic4cast}}$)} are notoriously difficult. For example, ``Despite the Porto dataset's original use as a standardized benchmark for open competition, design choices in subsequent work make cross-paper comparison difficult. Firstly, different papers often augment the dataset with their metadata not present in the original release, which may give some models an advantage over others independent of architecture or training design.''~\cite{tenzer_meta-learning_2022}

The remainder of this paper is structured as follows: 
Section~\ref{sec:usecases} presents the recent deep learning research structured by the main use cases, neural network architecture, and trajectory data granularity and representation. 
Section~\ref{sec:motivation} summarizes and discusses the motivations provided for the use of DL over classic ML and the benchmarks and metrics used to evaluate their performance. 
Section~\ref{sec:trends} extends the time frame of our analysis to further analyze the trends in this research area. Finally, Section~\ref{sec:conclusion} summarizes the findings and presents our conclusions. 
Since the terminology used in different publications is not necessarily consistent due to the large range of domains working on trajectory data analysis, we, define the most important terms and abbreviations in a glossary in the paper appendix.

\section{Representing Trajectory Data For Deep Learning Use Cases}
\label{sec:usecases}

Trajectory datasets used in research and industry are highly heterogeneous since they are affected by numerous technical and design choices. For example, in Graser et al.~\cite{werner_exploratory_2021}, we distinguish 10 dimensions along which movement dataset may vary:  
\begin{enumerate*}
    \item Spatial resolution (fine or coarse / small or large location errors), 
    \item Spatial dimensions (2D or 3D), 
    \item Temporal resolution (sparse or frequent / quasi-continuous),
    \item Sampling interval (regular or irregular),
    \item Representation (polylines or continuous curves),
    \item Constraints (network-constrained or unconstrained / open space),
    \item Collection models (Lagrangian or Eulerian / checkpoint-based),
    \item Tracking system (cooperative or uncooperative),
    \item Privacy (personal or impersonal),
    \item Data size (small or big datasets / streams).
\end{enumerate*}
It is therefore not hard to imagine that there can be no one-size-fits-all approach to modelling trajectory data for machine learning. Instead, it is necessary to pick the right model for the use case and data available.

The literature on deep learning from trajectory data can be roughly divided into eight use case categories, as shown in Table~\ref{table:overview}.
The first three use cases, consisting of:  
1. Trajectory prediction (\& imputation),  
2. Arrival time prediction mostly, and
3. (Sub)trajectory classification
mostly use dense / quasi-continuous trajectory data, while the following use cases:  
4. Anomaly detection,
5. Next location (\& destination) prediction,
6. Synthetic data generation,
7. Location classification use cases mostly, and
8. Traffic volume / crowd flow prediction 
tend to use increasingly sparse / check-in data or aggregated trajectories of multiple movers.

\begin{table*}[ht]
 \centering
 \scriptsize
 \caption{Overview of trajectory data representations used to train neural networks using the following dataset categorization: trajectories of individual movers marked with \color{blue}$\Rightarrow$ for dense / quasi-continuous, \color{violet}$\to$ for sparse/check-in data or gridded data \color{black} and \color{red}$\rightleftharpoons$ for aggregated trajectories of multiple movers\color{black}. References marked with asterisk (*) mark review / summary papers}
 \label{table:overview}
   \begin{tabularx}{1\textwidth}{ >{\hsize=55pt}X !{\color{lightgray}\vline} X !{\color{lightgray}\vline} X !{\color{lightgray}\vline} X !{\color{lightgray}\vline} X !{\color{lightgray}\vline} X !{\color{lightgray}\vline} X !{\color{lightgray}\vline} >{\hsize=60pt}X } 
      \toprule
    \textbf{Use case} & \textbf{CNN} & \textbf{(C)RNN} & \textbf{LSTM} & \textbf{GRU} & \textbf{Trans\newline former} & \textbf{GNN} & \textbf{Other} \\  
   \cmidrule{1-8}

\rowcolor{traj!10}\cellcolor{traj}\color{black}
Trajectory\newline prediction /\newline imputation \tiny{[\ref{uc:trajpred}]}
   &  &  & \color{blue}$\Rightarrow$ \cite{mehri_contextual_2021} \newline \color{blue}$\Rightarrow$ \cite{capobianco_recurrent_2023} & \color{blue}$\Rightarrow$ \cite{tritsarolis_online_2021} \newline \color{violet}$\to$
 \cite{fan_online_2022} & \color{violet}$\to$ \cite{carroll_towards_2022,musleh_towards_2022} & &  
\\ 
\rowcolor{arrival!10}\cellcolor{arrival}\color{white}
Arrival time\newline prediction \tiny{[\ref{uc:eta}]}
   &  
   &  \color{blue}$\Rightarrow$ \cite{wang_when_2018,yin_deep_2021}  &  & \color{blue}$\Rightarrow$ \cite{yin_deep_2021} 
   &  & \color{red}$\rightleftharpoons$ \cite{derrow-pinion_eta_2021} & 
\\  
\rowcolor{subtraj!10}\cellcolor{subtraj}\color{black}
(Sub)trajectory\newline classification \tiny{[\ref{uc:subtraj}]}
   & \color{violet}$\to$ \cite{chen_ship_2020,yang_ship_2022} & \color{blue}$\Rightarrow$ \cite{wang_locomotion_2021}* & \color{blue}$\Rightarrow$ \cite{wang_locomotion_2021}* &  & \color{blue}$\Rightarrow$ \cite{wang_locomotion_2021}*  & & \color{blue}$\Rightarrow$ AdaNet: \cite{wang_locomotion_2021}* 
\\  
\rowcolor{anomaly!10}\cellcolor{anomaly}\color{white}
Anomaly\newline detection \tiny{[\ref{uc:anomaly}]}
   &  & \color{violet}$\to$ \cite{nguyen_geotracknet_2022}
   \newline \color{red}$\rightleftharpoons$ \cite{singh_leveraging_2022} & \color{blue}$\Rightarrow$ \cite{liatsikou_denoising_2021} &  &  &  &  
\\ 
\rowcolor{nextloc!10}\cellcolor{nextloc}\color{white}
Next location /\newline destination\newline prediction \tiny{[\ref{uc:nextloc}]}
   & \color{violet}$\to$ \cite{gao_predicting_2019} & \color{violet}$\to$ \cite{feng_deepmove_2018,liao_predicting_2018} &  & \color{violet}$\to$ \cite{gao_predicting_2019} & \color{violet}$\to$ \cite{hong_how_2022} & & \color{blue}$\Rightarrow$ Hyper \newline network: \cite{tenzer_meta-learning_2022} \newline \color{violet}$\to$ SAN: \cite{li_predicting_2020}
\\  

\rowcolor{synth!10}\cellcolor{synth}\color{black}
Synthetic data\newline generation  \tiny{[\ref{uc:synth}]}
   & & & \color{violet}$\to$ \cite{rao_lstm-trajgan_2020} & & & 
   & \color{blue}$\Rightarrow$ VAE-like:  \cite{zhang_factorized_2022} \newline 
   \color{violet}$\to$ GAN: \cite{rao_lstm-trajgan_2020} \newline \color{red}$\rightleftharpoons$ MLP: \cite{simini_deep_2021} 
\\ 
\rowcolor{location!10}\cellcolor{location}\color{white}
  Location\newline classification  \tiny{[\ref{uc:location}]}
    & \color{violet}$\to$ \cite{yang_classification_2018} 
    &  &  &  &  
    & \color{red}$\rightleftharpoons$ \cite{altan_discovering_2022} 
    & \color{violet}$\to$ Memory \newline network: \cite{lyu_plug-memory_2022} 
\\ 
\rowcolor{volume!10}\cellcolor{volume}\color{white} 
Traffic volume\newline / crowd flow\newline prediction \tiny{[\ref{uc:volume}]}
   & \color{red}$\rightleftharpoons$ \cite{wang_deep_2022,lu_learning_2021,kashyap_traffic_2022} & \color{red}$\rightleftharpoons$ \cite{kashyap_traffic_2022} & \color{red}$\rightleftharpoons$ \cite{buroni_tutorial_2021} \newline  \color{violet}$\to$ \cite{li_prediction_2021} &  & \color{red}$\rightleftharpoons$ \cite{xue_leveraging_2022} &  \color{violet}$\to$ \cite{gao_traffic_2022} \newline \color{red}$\rightleftharpoons$ \cite{li_prediction_2021,lippert_learning_2022} &  \color{red}$\rightleftharpoons$ GAN: \cite{zhang_curb-gan_2020} \newline \color{red}$\rightleftharpoons$ SAE:  \cite{kashyap_traffic_2022} 
   \\ \hline

 \end{tabularx}
\end{table*}

\begin{figure}[ht!]
  \centering
  \includegraphics[width=0.69\linewidth]{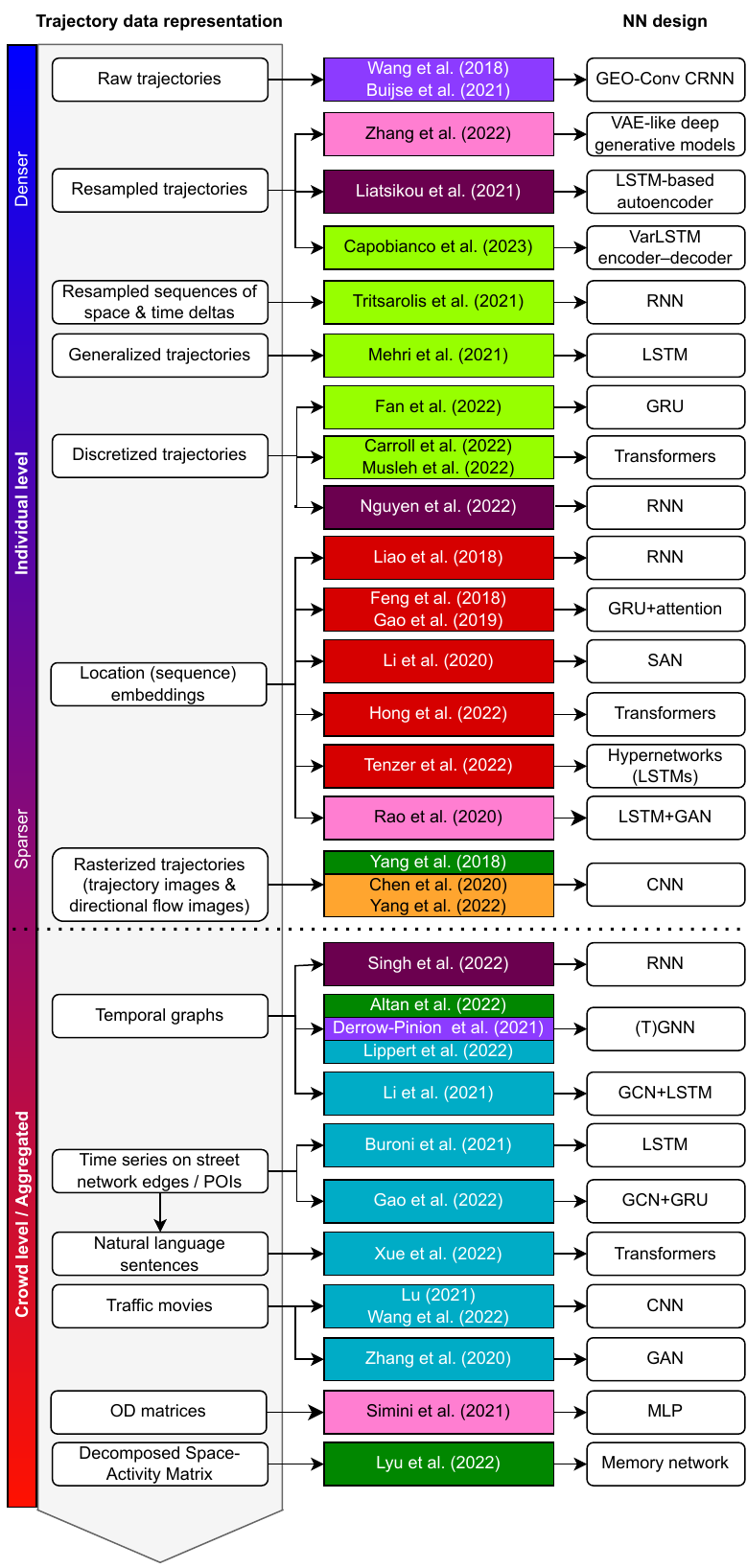}
  \caption{Overview of trajectory data representations used to train neural networks}
  \label{fig:overview}
\end{figure}

To further illustrate the wide variety of different trajectory data representations encountered in these use cases, 
Figure~\ref{fig:overview} provides an overview ordered by the identified trajectory representations from dense individual trajectories at the top to crowd-level / aggregated movement trajectory data at the bottom. 
In many instances, the borders between these categories are fuzzy since they reside on a spectrum. 
While use cases such as \textit{Trajectory prediction}, \textit{Next location prediction} and \textit{Traffic volume / crowd flow prediction} are clearly clustered in certain sections along this spectrum, other use cases, such as \textit{Anomaly detection} and \textit{Synthetic data generation} appear all over since, for example, anomalies can be analyzed both on the level of individual trajectories as well as on the crowd level where analysts may be interested in detecting unusual crowd patterns.

More details on the trajectory datasets and the data engineering steps applied to the trajectory data before they are used as input to train the neural networks are summarized in Tables~\ref{table:dataengineering}-\ref{table:dataengineering3}.
Some works (e.g. \cite{gao_traffic_2022,rao_privacypreserving_2021}) use additional data sources (such as the road network and POIs) in combination with trajectory data to train their models. These additional data sources have not been included in our review in favor of clarity and conciseness. 

The following subsections describe each use case in more detail and introduce the neural network designs used to address the use cases, as well as the trajectory data representations used to train these neural networks.

\subsection{Trajectory prediction / imputation}
\label{uc:trajpred}

This use case category covers the prediction of trajectories in artificial  \cite{carroll_towards_2022},
urban \cite{fan_online_2022}, and maritime environments \cite{mehri_contextual_2021,tritsarolis_online_2021,capobianco_recurrent_2023}, as well as the 
imputation of trajectories \cite{musleh_towards_2022}. 
Depending on the context, trajectory prediction is also called movement prediction~\cite{mehri_contextual_2021,tritsarolis_online_2021} or future location prediction (FLP)~\cite{tritsarolis_online_2021} which can lead to confusions with the next location / destination prediction use case. However, this use case differs from the next location / destination prediction use case (which we will discuss later) in that it aims to predict the future path of a mover, that is, along which path the mover will get to its next location.

To predict trajectory paths with high spatial detail, it is necessary to learn from high-resolution training data. Therefore, the training data for these models consists of rather high-density trajectories of individual movers, which may be resampled, generalized, or discretized, as shown in Figure~\ref{fig:overview}. For example, \added{Capobianco et al.~\cite{capobianco_recurrent_2023} resample each trajectory to a fixed sampling interval of 15 minutes and create fixed-length trajectories with 12 positions, equalling a fixed trajectory duration of three hours.} 
Mehri et al.~\cite{mehri_contextual_2021}  generalize AIS trajectories using context-aware piece-wise linear segmentation before feeding them into an LSTM three vertices at a time. This enables their model to perform short-term trajectory predictions with high spatial detail. 
Tritsarolis et al.~\cite{tritsarolis_online_2021}, on the other hand, represent trajectories by their composition of differences in space \(\Delta x\), \(\Delta y\), and time \(\Delta t\) for the input of their RNN-based models to predict future positions \(\Delta t+1\). They even take the problem one step further and try to predict the co-movement of multiple movers through evolving clusters. 

An example of more long-term trajectory predictions is presented in Fan et al.~\cite{fan_online_2022}. They discretize mobile phone GPS trajectories using the H3 hexagonal grid~\footnote{\url{https://github.com/uber/h3}} and use the grid cell sequences to train their GRU. The resulting model predicts cell sequences which are afterwards used to search for similar high-resolution trajectories, which are returned as the final trajectory prediction.

Transformers are another approach gaining traction. For example, Carroll et al.~\cite{carroll_towards_2022} use synthetic discrete movement sequences in a minimalistic grid world environment to train their transformers to predict trajectories. 
Another work using transformers to impute trajectories is Musleh et al.~\cite{musleh_towards_2022}. They propose TrajBERT, a model trained using H3-discretized (tokenized) GPS tracks. They ``down-sample the trajectories by dropping three-quarters of the points of each trajectory and then run TrajBERT to fill the gaps by imputing the missing points''.

The difficulty of a specific trajectory prediction task is influenced by the type of movement (network-constrained or unconstrained), the complexity of the movement patterns observed in the area of interest \cite{graser_data-driven_2019}, as well as the desired prediction horizon (how far into the future the prediction is attempted) and the quality of the available training data. 
\added{Additional information, such as the vessel destination labels~\cite{capobianco_recurrent_2023} can further improve the prediction accuracy.}
The evaluation metrics for trajectory prediction tasks usually include the RMSE, MAE, or MAPE of spatial and spatiotemporal distance measures (see Table~\ref{table:comparisons}).

\subsection{Arrival time prediction} 
\label{uc:eta}

This use case category covers the prediction of travel times or arrival times, such as arrival time prediction in train networks \cite{yin_deep_2021} and 
street networks \cite{derrow-pinion_eta_2021,wang_when_2018}. This type  of task is also known as Estimated Time of Arrival (ETA) task. In classical intelligent transportation systems (ITS), ETA is commonly computed using routing algorithms leveraging street network data with travel time information. This travel time information may be historical averages for a certain season, day of the week, and time of day, or the result of more advanced ML/DL prediction models. 
Since future travel times often correlate with historical travel times (for example, at a given time of day and day of the week), recurrent mechanisms are commonly used to predict them \cite{derrow-pinion_eta_2021,wang_when_2018,yin_deep_2021}. For example, Derrow-Pinion et al.~\cite{derrow-pinion_eta_2021} train GNNs on aggregated trajectories to provide travel time predictions in Google Maps. The GNN graph consists of segment and supersegment-level embedding vectors. Nodes store street segment-level data (average real-time and historical segment travel speeds and times, segment length, and road class), while edges store supersegment-level data (real-time supersegment travel times).

In contrast, Wang et al.~\cite{wang_when_2018} introduces the GEO-convolutional network layer (\textit{GEO-Conv}, also used by Buijse et la.~\cite{yin_deep_2021}), which is trained on dense trajectories rather than aggregated trajectories. 
The proposed \textit{GEO-Conv layer} takes dense trajectories  as input and applies a non-linear mapping of each trajectory point, followed by a GEO-Conv step with multiple kernels. 

The difficulty of a specific arrival time prediction task is influenced by the variability of the travel times in the area of interest, as well as the desired prediction horizon  and the quality of the available training data (including sufficient data on relevant recurring patterns and events that may influence travel times). 
The evaluation metrics for arrival time prediction tasks usually include the RMSE, MAE and MAPE (see Table~\ref{table:comparisons}).

\subsection{(Sub)trajectory classification}
\label{uc:subtraj}

This use case category considers the classification of complete trajectories \cite{yang_ship_2022} or sub-trajectories \cite{chen_ship_2020,wang_locomotion_2021}. 
Typical applications include the detection of mover classes and movement types, such as ship maneuvers \cite{chen_ship_2020} or the detection of transportation and locomotion modes of smartphone users \cite{wang_locomotion_2021}. 

An example of the classification of complete trajectories is the recognition of ship types introduced by Yang et al.~\cite{yang_ship_2022}. It relies on the same technique as \cite{chen_ship_2020} for transforming the raw AIS trajectory data into colour-coded trajectory images. The resulting images show characteristic trajectory patterns, which can be used to identify the ship vessel type with a CNN classifier.

By splitting trajectories into sub-trajectories, more fine-grained analyses are possible. 
For example, Chen et al.~\cite{chen_ship_2020} generate colour-coded trajectory images from ship AIS data, where each pixel is assigned one of three colors according to the movement type (static, normal, maneuvering). These trajectory images are used to train a CNN-based ship maneuver classifier. 

An example for human sub-trajectory classification is presented in the Sussex-Huawei Locomotion-Transportation (SHL) Recognition Challenge~\cite{wang_locomotion_2021} which aims to identify different movement modes (i.e. still, walk, run, bike, car, bus, train, and subway) from smartphone data. The SHL Challenge winner uses AdaNet, a TensorFlow-based framework for learning NN models and ensembling models to obtain even better models~ \cite{cortes_adanet_2017}.

The difficulty of a specific (sub)trajectory classification task depends on the the nature of the classes of interest (some classes may easy to distinguish while others may have large overlaps with neighboring classes) as well as the quality of the training dataset (in particular the class balance which may be more or less skewed towards the popular classes and under-represent rarer classes). Additional complexity is introduced if -- in addition to the classification -- the splits between subtrajectories have to be determined as well.
The evaluation metrics for (sub)trajectory classification tasks usually include F1 score, accuracy, precision and recall (see Table~\ref{table:comparisons}).

\subsection{Anomaly detection}
\label{uc:anomaly}

This use case category deals with identification and handling of unusual observations and patterns in the data, often referred to as outliers or anomalies. Both terms are often used interchangeably~\cite{chandola_anomaly_2009}. Outliers can be defined as observations that ``deviate so much from other observations as to arouse suspicions that it was generated by a different mechanism''~\cite{hawkins_identification_1980}. Similarly, anomalies are patterns that do not conform to an expected behavior~\cite{chandola_anomaly_2009}. 

Types of movement anomalies include anomalous records (unusual spatiotemporal or thematic attributes), anomalous (sub)trajectories, and anomalous events (when ``the trajectories of individual movers are unremarkable but their combined spatio-temporal pattern is unusual''~\cite{werner_exploratory_2021}).
Since the definition of anomalies is often context-dependent, ground truth labeled data is rare. Therefore, anomaly detection approaches often resort to trying to identify trajectories that deviate significantly from previously observed trajectories based on some spatial, spatiotemporal, or other metrics. 
GeoTrackNet~\cite{nguyen_geotracknet_2022}, for example, is a model for maritime trajectory anomaly detection, which consists of a probabilistic RNN-based (Recurrent Neural Network) representation of AIS tracks and a contrario detection~\cite{desolneux_gestalt_2007}. Anomalies detected by GeoTrackNet were then evaluated by AIS experts. Similarly, Singh et al.~\cite{singh_leveraging_2022} present an anomaly detection system based on RNN regression models to detect anomalous trajectories, on-off switching, and unusual turns. Again, a quantitative accuracy analysis is not feasible due to the lack of ground truth data.

To address the issue of lacking ground truth, some researchers resort to using synthetically generated anomalies~\cite{liatsikou_denoising_2021}. For example, Liatsikou et al.~\cite{liatsikou_denoising_2021} developed an LSTM-based network for the automatic detection of movement anomalies, such as the detection of synthetic anomalies in taxi trajectories. The anomalies that can be detected are of limited length since the autoencoder requires inputs of a certain fixed length, all trajectories are clipped to nine points (and shorter ones discarded). 

The difficulty of a specific anomaly detection task depends on the anomalies of interest (and how much they deviate from regularly observed behavior) as well as the availability of labelled training data. In settings where only detected anomalies can be presented to experts for assessment, there is a lack of dependable quantitative data on true and false negatives. 
The evaluation metrics for anomaly detection tasks usually include accuracy, precision and recall (see Table~\ref{table:comparisons}).

\subsection{Next location / destination prediction}
\label{uc:nextloc}

This use case category covers the prediction of the next locations or final destinations of trips \cite{feng_deepmove_2018,gao_predicting_2019,hong_how_2022,li_predicting_2020,liao_predicting_2018,liu_predicting_2016,tenzer_meta-learning_2022}.
This usually boils down to predicting the next location or final destination from a finite set of potential locations. 
Besides GPS tracks, a commonly used data source in this category are social media check-ins (e.g., from Foursquare). The task then becomes to predict the next check-in location.  

Attention mechanisms are particularly popular for next location prediction.
For example, Gao et al.~\cite{gao_predicting_2019} train \textit{VANext}, a semi-supervised network trajectory convolutional network, on check-in data. They convert each individual user's trajectory (check-in/POI sequence) into sequence embeddings using a causal embedding method (similar to a high-order Markov process). The resulting embeddings are the input for their GRU to learn the trajectory patterns. They further apply attention to the embeddings for predicting the user's next POI.
Feng et al.~\cite{feng_deepmove_2018} tailor two attention mechanisms to generate independent latent vectors from large and sparse trajectories. These embeddings are then fed into their \textit{DeepMove} GRUs and a historical attention module. The learned attention weights can intuitively explain the prediction based on the user's history of movement behavior.
Li et al.~\cite{li_predicting_2020} introduce a spatio-temporal self-attention network (\textit{STSAN}). They generate trajectory embeddings by concatenating the temporal (activity sequence), spatial (distance matrix of locations), and location attentions (location sequence and their categories). They feed these embeddings through a softmax layer and predict the user's next POI. They use a federated learning setting to tackle the heterogeneity problem. 
Liao et al.~\cite{liao_predicting_2018} generate embeddings from location sequences as well as graph embeddings from location-location and activity-location graphs and train their \textit{MCARNN} multi-task context-aware recurrent neural network to solve both activity and location prediction tasks.

Other works use neural networks for dimensionality reduction and for creating embeddings.
Liu et al.~\cite{liu_predicting_2016} incorporate time and distance-specific transition matrices as temporal and spatial embeddings generated by RNNs.
Hong et al.~\cite{hong_how_2022} reduce the dimensions of trajectories using a multilayered embedding approach for transformers to predict next location and travel mode. 
Tenzer et al.~\cite{tenzer_meta-learning_2022} generate two geospatial and temporal embeddings by 
\begin{enumerate*}
    \item combining the random picking and the nearest neighbor to create sequences of spatial embeddings and
    \item using a sinusoidal embedding to convert the timesteps to temporal vectors.
\end{enumerate*}
They train a hyper network to learn to change its weights in response to these embeddings.

The difficulty of a specific next location / destination prediction task depends on many of the same factors discussed for trajectory prediction. 
The evaluation metrics for next location / destination prediction tasks usually include accuracy, precision and recall which pick up on the fact that predicting the location from a finite set of potential locations is reminiscent of a classification task. An alternative approach is to treat the task like a regression task and measure, for example, the RMSE of the spatial distance between the predicted and the true location (see Table~\ref{table:comparisons}).

\subsection{Synthetic data generation}
\label{uc:synth}

This category covers the generation of synthetic movement data, such as synthetic trajectories \cite{zhang_factorized_2022,rao_lstm-trajgan_2020} and synthetic flows \cite{simini_deep_2021}. The creation of synthetic data is of particular interest to either address data privacy concerns or to deal with a lack of real data for certain regions. 
For example, Rao et al.~\cite{rao_lstm-trajgan_2020} address the privacy issue by developing an end-to-end deep LSTM-TrajGAN model to generate privacy-preserving synthetic trajectory data for data sharing and publication. Similarly, 
Zhang et al.~\cite{zhang_factorized_2022} propose an end-to-end trajectory generation model for generating synthetic trajectories using VAE-like encoders 
and decoders where a prior generator based on variational recurrent structure generates noise at time $t$ by considering the noise at the previous time step.  

Aggregated flow data, on the other hand is usually not privacy sensitive but it may not be available in some regions or for certain time periods. Therefore, Simini et al.~\cite{simini_deep_2021} developed an MLP model (denoted Deep Gravity) to generate mobility ﬂow probabilities. They evaluated Deep Gravity on mobility ﬂows in England, Italy, and New York State and achieved a good performance even for regions with no data available for training.

The difficulty of a synthetic data generation task, as well as the selection of suitable evaluation metrics for synthetic data generation depend on the specific details of the application (for example, whether it aims at generating individual trajectories or aggregated flows) and method used (see Table~\ref{table:comparisons2}). 

\subsection{Location classification}
\label{uc:location}

This use case category covers the classification of locations, such as certain areas or points of interest (POIs), using patterns derived from movement data. The classification of regionally dominant movement patterns may be of interest in and of itself~\cite{yang_classification_2018} or it may help with the classification of POIs (e.g. ports~\cite{altan_discovering_2022}) and trip destinations~\cite{lyu_plug-memory_2022}.

To detect regionally dominant movement patterns, Yang et al.~\cite{yang_classification_2018} use direction information and density maps to generate directional flow images. They convert trajectories into images where each pixel contains the directional flows. They use a CNN to classify the input image patterns and detect the dominant regional movements. 

An approach that makes more use of the temporal information contained in trajectories is presented by Altan et al.~\cite{altan_discovering_2022}. They use a temporal GNN (TGNN) to distinguish gateway ports from actual ports using AIS vessel movement data. After extracting the ports (nodes) from the raw AIS messages using DBSCAN, they extract trips between consecutive ports and build a graph for each time step to generate the time-ordered daily graph sequence for the TGNN. 

Lyu et al.~\cite{lyu_plug-memory_2022} train a model to predict trip purposes based on destination locations. Their model is trained using activity, origin, and destination matrices derived from OD data. 
They use latent mode alignment to align the geographic contextual latent with travel activities. This approach is called the plug-in memory network.

The difficulty of a specific location classification task depends on the nature of the location classes of interest and other factors previously discussed for (sub)trajectory classification tasks. An additional factor is whether the evaluation is performed using locations that fall into the same geographic region as the training locations or whether the the evaluation locations are from a different region (thus adding the challenge of model transferability to different geographic contexts).
The evaluation metrics for location classification usually include precision and recall (see Table~\ref{table:comparisons2}).

\subsection{Traffic volume prediction}
\label{uc:volume}

This use case category covers the prediction of traffic or crowd volumes and flows, including, for example, predicting traffic volume on street segments \cite{buroni_tutorial_2021,kashyap_traffic_2022},
human activity at specific POIs \cite{gao_traffic_2022} and metropolitan areas \cite{li_prediction_2021,lu_learning_2021}, or predicting animal movement dynamics \cite{lippert_learning_2022}.

Models for this use case are commonly trained with aggregated trajectory data. For example, traffic movies are the training data provided for the \textit{Traffic4Cast 2021} competition which challenged participants to  predict traffic under conditions of temporal domain shift (Covid-19 pandemic) and spatial shift (transfer to entirely new cities). Lu~\cite{lu_learning_2021} won this challenge using CNN (U-Net) and multi-task learning. Their multi-task learning approach randomly samples from all available cities and trains the U-Net model to jointly predict the future traffic states for different cities. 
Wang et al.~\cite{wang_periodic_2022} also use the traffic movie approach. They aggregate individual-level trajectories into a grid with inflow referring to the total number of incoming traffic entering this region from other regions during a given time interval and outflow representing the total number of traffic leaving the region.
Similarly, Zhang et al.~\cite{zhang_curb-gan_2020} create temporal grids of average traffic speed and taxi inflow per cell.

Graph-based models represent another common approach for this use case. For example, Li et al.~\cite{li_prediction_2021} build a graph for their GCN by aggregating CDR data and representing spatial statistical units as nodes and their relationship (physical distance, physical movement, phone calls) as edges.
Similarly, Lippert et al.~\cite{lippert_learning_2022} build temporal graphs from bird migration data where nodes represent radar locations, and edges represent the flows between the Voronoi tessellation cells of the radar locations. 

Other approaches include, for example, Buroni et al.~\cite{buroni_tutorial_2021} who provide a  tutorial using vehicle counts derived from GPS tracks to build and train a Direct LSTM encoder-decoder model. The model is trained to predict counts of vehicles per network edge per time step for the Belgian motorway network.
Similarly, Gao et al.~\cite{gao_traffic_2022} use their GPS tracks to count vehicles per POI per time step (hourly) to train a GCN+GRU model that predicts these visit counts.
And Xue et al.~\cite{xue_leveraging_2022}  propose a translator called \textit{mobility prompting} which converts daily POI visit counts into natural language sentences so they can use (and fine-tune) pre-trained NLP models such as Bert, RoBERTa, GPT-2, and XLNet to predict these visit counts. 

The difficulty of a specific traffic volume prediction task is influenced by the variability of the traffic volumes in the area of interest, as well as the desired prediction horizon and the quality of available training data.  
The evaluation metrics for traffic volume or crowd flow prediction usually include RMSE, MAE, and MAPE (see Table~\ref{table:comparisons2}).

\begin{landscape}

\begin{table*}[ht]
 \centering
 \scriptsize
 \caption{Trajectory datasets and data engineering; public datasets are printed in bold}
 \label{table:dataengineering}
   \begin{tabularx}{1.55\textwidth}{ 
       >{\hsize=.20\hsize}X !{\color{lightgray}\vline} 
       >{\hsize=.22\hsize}X !{\color{lightgray}\vline} 
       X  !{\color{lightgray}\vline} 
       X } 
   \hline
    \textbf{Ref} & \textbf{NN design} & \textbf{Trajectory data} & \textbf{Data engineering} \\ 
   \rowcolor{traj}\multicolumn{4}{l}{\color{black}
   \textbf{Trajectory prediction / imputation}} 
   \\  
   Mehri et al. (2021) \cite{mehri_contextual_2021}  
   & LSTM
   & \textbf{AIS data from NOAA}\footnotemark \space for the US East Coast, containing 58.5mio messages from 10.7k vessels over 2 months 
   & Trajectory is generalized using context-aware piecewise linear segmentation. The LSTM is trained on three vertices at a time 
   \\  \arrayrulecolor{lightgray}  \cmidrule{1-4}  
   Tritsarolis et al. (2021) \cite{tritsarolis_online_2021}
   & RNN
   & \textbf{Ship AIS tracks in Piraeus}\footnotemark \space containing 138k records from 246 fishing vessels   and \textbf{GeoLife}$^{\ref{footnote-geolife}}$ \space
   & Trajectories are resampled to regular 1~minute intervals and converted into sequences of differences in space \(\Delta x\), \(\Delta y\), and time \(\Delta t\) to predict the vessel's position in the next future step
   \\  \arrayrulecolor{lightgray}  \cmidrule{1-4}  
   Fan et al. (2022) \cite{fan_online_2022}  
   & GRU 
   & Mobile phone GPS tracks in the Kanto area of Tokyo covering 220k users with a minimum reporting period of 5 minutes for 2 months
   & Trajectories are \textbf{discretized using the H3} hexagonal grid
   \\  \arrayrulecolor{lightgray}  \cmidrule{1-4}  
   Carroll et al. (2022) \cite{carroll_towards_2022}  
   & Transformers
   & Synthetic data in a minimalistic grid world environment 
   & Transformers are trained on  trajectories as sequences of states, actions, and return-to-go tokens
   \\  \arrayrulecolor{lightgray}  \cmidrule{1-4}    
   Musleh et al. (2022) \cite{musleh_towards_2022} 
   & Transformers 
   & GPS tracks in San Francisco from the GISCUP’17 dataset with 5M records 
   & Trajectories are \textbf{discretized using the H3} hexagonal grid (tokenization) followed by the creation of spatial embeddings
   \\  \arrayrulecolor{lightgray}  \cmidrule{1-4}    
   \added{Capobianco et al. (2023) \cite{capobianco_recurrent_2023}}
   & \added{LSTM}
   & \added{\textbf{AIS data from the Danish Maritime Authority}}\footnotemark, \added{394 trajectories of tanker vessels belonging to two specific motion patterns of interest over one month}
   & \added{Trajectories are resampled to a fixed sampling interval of 15 minutes and segmented into fixed-length trajectories with 12 positions}
   \\  
   \rowcolor{arrival}\multicolumn{4}{l}{\color{white}\textbf{Arrival time prediction}} 
   \\
   Wang et al. (2018) \cite{wang_when_2018} 
   & CRNN 
   & Taxi GPS tracks (FCD) in Chengdu and Beijing with a density of 2.6 to 5.5 GPS records per km
   & \textbf{Trajectories} are fed into a \textit{GEO-Conv} layer
   \\  \arrayrulecolor{lightgray}  \cmidrule{1-4}
   Buijse et la. (2021) \cite{yin_deep_2021} 
   & CRNN+GRU 
   & Train GPS tracks in the Netherlands covering 350k train trips  
   & \textbf{Trajectories} are fed into a \textit{GEO-Conv} layer
   \\  \arrayrulecolor{lightgray}  \cmidrule{1-4}
   Derrow-Pinion et al. (2021) \cite{derrow-pinion_eta_2021} 
   & GNN 
   & Google Maps road segments with travel times/speeds 
   & Road network graph edges are subdivided into shorter segments (modeled as \textbf{GNN graph} nodes) with associated aggregated travel time/speed information; segments are combined into supersegments (GNN graph edges)
    \\ 

   \rowcolor{subtraj}\multicolumn{4}{l}{\color{black}\textbf{(Sub)trajectory classification}} 
   \\
   Chen et al. (2020) \cite{chen_ship_2020} 
   & CNN 
   & \textbf{Ship AIS tracks in Tianjin, China} covering 23k trips
   & Trajectories are converted into \textbf{trajectory images} with pixels coloured according to movement type (static, maneuvering, normal)
   \\  \arrayrulecolor{lightgray}  \cmidrule{1-4}  
   Yang et al. (2022) \cite{yang_ship_2022} 
   & CNN 
   &  \textbf{Ship AIS tracks in Northern America} of the U.S. National Oceanic and Atmospheric Administration’s Office of Coastal Management with 259k records after pre-processing 
   & Trajectories are converted into \textbf{trajectory images} with the same method as described in \cite{chen_ship_2020}. 
   \\  \arrayrulecolor{lightgray}  \cmidrule{1-4}  
   Wang et al. (2021) \cite{wang_locomotion_2021} 
   & (C)RNN, LSTM, Transformers, AdaNet 
   & SHL dataset\footnotemark: asynchronously sampled radio data of smartphones with up to 2,812 hours of labeled data, e.g. GPS reception and location, Wifi reception and GSM cell tower scans 
   & [SHL challenge summary paper] Mostly hand-crafted feature engineering as input, but also two teams using raw trajectory data 
   \\ 

   \\  \arrayrulecolor{black}  \bottomrule
 \end{tabularx}
\end{table*}

\begin{table*}[ht]
 \centering
 \scriptsize
 \caption{Trajectory datasets and data engineering; public datasets are printed in bold -- continued}
 \label{table:dataengineering3}
   \begin{tabularx}{1.55\textwidth}{ >{\hsize=.20\hsize}X !{\color{lightgray}\vline} >{\hsize=.22\hsize}X !{\color{lightgray}\vline} X  !{\color{lightgray}\vline} X } 
    \hline
    \textbf{Ref} & \textbf{NN design} & \textbf{Trajectory data} & \textbf{Data engineering} \\

   \rowcolor{anomaly}\multicolumn{4}{l}{\color{white}\textbf{Anomaly detection}} 
   \\
   Liatsikou et al. (2021) \cite{liatsikou_denoising_2021} 
   & LSTM-based AE 
   & \textbf{Porto taxi tracks}$^{\ref{footnote-porto}}$ (FCD) 
   & Trajectories are down-sampled to 60s and represented as a \textbf{sequence of vectors} (lat, lon) and clipped to the first nine points to fit the autoencoder requirements 
   \\  \arrayrulecolor{lightgray}  \cmidrule{1-4}  
   Nguyen et al. (2022) \cite{nguyen_geotracknet_2022} 
   &  {Probabilistic RNN} 
   & Ship AIS tracks from a single receiver in Ushant, France
   & Trajectories are down-sampled to 600s  and converted to a \textbf{four hot vectors} (lat, lon, SOG, COG) with the resolutions of 0.01° for lat/lon, 1 knot for SOG, and 5° for COG.
   \\  \arrayrulecolor{lightgray}  \cmidrule{1-4}  
   Singh et al. (2022) \cite{singh_leveraging_2022} 
   & RNN regression model 
   & Ship AIS tracks in the Baltic sea region and near Bremerhaven, Germany for two months (including a comparison between satellite-based and coastal AIS) 
   & Trajectories are resampled and interpolated at the 60s and converted into a \textbf{graph} with nodes representing turning points for the vessel trajectories and edges representing the sea lanes traveled by vessels
   \\ 

   \rowcolor{nextloc}\multicolumn{4}{l}{\color{white}\textbf{Next location / destination prediction}} 
   \\
   Gao et al. (2019) \cite{gao_predicting_2019}
   & GRU \& CNN
   & Foursquare check-ins in New York and Singapore and Gowalla data in Houston and California with an average of 229k records of 3k users 
   & Location (<id, lon, lat>) sequences are converted into \textbf{sequences embeddings} using a causal embedding method 
   \\  \arrayrulecolor{lightgray}  \cmidrule{1-4}  
   Feng et al. (2018) \cite{feng_deepmove_2018}
   & Attention-GRU 
   & \textbf{Foursquare check-ins}\footnotemark \space  with 294k records covering 15k users, other mobile application location records (search  \& check-ins) with 15mio records covering 5k users, and CDR with 491k records covering 1k users  
   & Location sequences are converted into \textbf{embeddings} using two independent attention mechanisms 
   which are then fed into \textit{DeepMove's} GRUs and a historical attention module
   \\  \arrayrulecolor{lightgray}  \cmidrule{1-4}  
   Li et al. (2020) \cite{li_predicting_2020}
   & SAN 
   & \textbf{Foursquare check-ins}, Tweets, Yelp and NYC data
   & Location sequences are converted into \textbf{embeddings} 
   \\  \arrayrulecolor{lightgray}  \cmidrule{1-4}   
   Liao et al. (2018) \cite{liao_predicting_2018}
   &  RNN 
   & \textbf{Foursquare check-ins} in NYC and Tokyo with an average of 400k records of 1.7k users
   & Location sequences and activity/location graphs are converted into \textbf{embeddings} for \textit{MCARNN}
   \\  \arrayrulecolor{lightgray}  \cmidrule{1-4}   
   Liu et al. (2016) \cite{liu_predicting_2016}
   & RNN
   & \textbf{Gowalla check-ins}$^{\ref{footnote-gowalla}}$  and \textbf{Global Terrorism Database (GTD) incidents}\footnotemark \space 
   & Location sequences are converted into time-specific and distance-specific transition matrices for \textit{ST-RNN} 
   \\
   Hong et al. (2022) \cite{hong_how_2022}
   & Transformers
   & Mobile phone GPS tracks in Switzerland from the Green Class (GC) study covering 139 participants for a year and from the yumuv study covering 498 participants for 3 months
   & Location sequences are generated from GPS tracks by first filtering stay locations with a stay duration >25min and then spatially aggregating stays into locations. These location sequences are converted into location, time, day, and mode embeddings which are fed into the transformers 
   \\  \arrayrulecolor{lightgray}  \cmidrule{1-4}   
   Tenzer et al. (2022) \cite{tenzer_meta-learning_2022}
   & Hyper network (LSTMs) 
   & \textbf{Porto taxi tracks}$^{\ref{footnote-porto}}$   
   covering 1.7mio trips by 442 taxis in Porto, Portugal for 12 months
   & Sequences of trajectory points are converted to \textbf{sequences of spatial embeddings} via a geospatial encoding mechanism 
\\  
\arrayrulecolor{black} 
   \bottomrule
   
 \end{tabularx}
\end{table*}

\begin{table*}[ht]
 \centering
 \scriptsize
 \caption{Trajectory datasets and data engineering; public datasets are printed in bold -- continued}
 \label{table:dataengineering2}
   \begin{tabularx}{1.55\textwidth}{ >{\hsize=.20\hsize}X !{\color{lightgray}\vline} >{\hsize=.22\hsize}X !{\color{lightgray}\vline} X  !{\color{lightgray}\vline} X } 
    \hline
    \textbf{Ref} & \textbf{NN design} & \textbf{Trajectory data} & \textbf{Data engineering} \\

   \rowcolor{synth}\multicolumn{4}{l}{\color{black}\textbf{Synthetic data generation}} 
   \\
   Rao et al. (2020) \cite{rao_lstm-trajgan_2020} 
   & LSTM-GAN 
   & \textbf{Foursquare check-ins}\footnotemark \space  in NYC covering 193 users with 3k trajectories and 67k records
   & Trajectories are processed by a trajectory encoding model covering trajectory point encodings (location, temporal and categorical attributes) and trajectory padding (to ensure that all trajectories have the same length)
   \\  \arrayrulecolor{lightgray}  \cmidrule{1-4}   
   Zhang et al. (2022) \cite{zhang_factorized_2022} 
   & VAE-like deep generative models 
   & \textbf{Porto taxi tracks}$^{\ref{footnote-porto}}$ (FCD); \textbf{T-Drive}$^{\ref{footnote-tdrive}}$ data consisting of 10.3k taxis for one week; and \textbf{Gowalla check-ins}$^{\ref{footnote-gowalla}}$ 
   & A coordinate encoding MLP converts two-dimensional points into a high-dimensional representation. Then, a Bidirectional LSTM is used to encode all representations with forward and backward information for a time step
   \\  \arrayrulecolor{lightgray}  \cmidrule{1-4}    
   Simini et al. (2021) \cite{simini_deep_2021} 
   & MLP 
   & England \& Italy commuting flows, \textbf{NY State flows} including  origin \& destination geographic unit and estimated population flows between two geographic units
   & Input to the model is of the origin \&  destination location as well as the distance between origin and destination. The output of the model is the probability to observe a trip between two locations.   \\    
    
   \rowcolor{location}\multicolumn{4}{l}{\color{white}
   \textbf{Location classification}} 
   \\  
   Yang et al. (2018) \cite{yang_classification_2018} 
   & CNN 
   & Synthetic data (manually drawn trajectories, rotated in a data augmentation step) 
   & Trajectories are converted to \textbf{directional flow images (DFI)} (resolution: 10×10)
   \\ \arrayrulecolor{lightgray}  \cmidrule{1-4}
   Altan et al. (2022) \cite{altan_discovering_2022} 
   & TGNN 
   & Ship AIS tracks in Halifax, Canada, covering 15 ports, 10 vessels, for 4 months (total: 513k AIS records) 
   & Trajectories are converted to \textbf{temporal (daily) graphs} of ports (nodes with associated port visit frequencies, waiting times, and speed statistics) and trips between consecutive ports (edges)
   \\  \arrayrulecolor{lightgray}  \cmidrule{1-4}
   Lyu et al. (2022) \cite{lyu_plug-memory_2022} 
   & Memory network
   & OD data from Tokyo Metro travel survey including travel mode, time, and purpose 
   & ODs are converted into a \textbf{Space-Activity Matrix} which is then decomposed into an activity, an origin, and a destination matrix
   \\ 

   \rowcolor{volume}\multicolumn{4}{l}{\color{black}\textbf{Traffic volume prediction}} 
   \\
   Lu (2021) \cite{lu_learning_2021} 
   & CNN (U\nobreakdash-Net) 
   & Traffic movies for 10 cities in 2019+2020 with 8 dynamic channels encoding traffic speed and volume per direction and 9 static channels encoding the properties of the road maps 
   & The multi-task learning randomly samples from all available city \textbf{traffic movies} (resolution: 495×436)
   \\  \arrayrulecolor{lightgray}  \cmidrule{1-4}
   Wang et al. (2022) \cite{wang_periodic_2022} 
   & CNN 
   & Taxi GPS tracks (FCD) from TaxiBJ in Beijing for 17 months and bike GPS tracks from BikeNYC in New York City for 6 months 
   & Trajectories are converted to \textbf{flow/traffic movies} (32×32 for TaxiBJ \& 16×8 for BikeNYC) with two dynamic channels encoding inflows and outflows
   \\  \arrayrulecolor{lightgray}  \cmidrule{1-4}
   Li et al. (2021) \cite{li_prediction_2021} 
   & GCN+LSTM 
   & Sparse CDR trajectories in Senegal of 100,000 individuals for one year  
   & Trajectories are converted into a \textbf{movement graph} with edges representing the number of transitions from one cell phone tower node to the next
   \\  \arrayrulecolor{lightgray}  \cmidrule{1-4}
   Buroni et al. (2021) \cite{buroni_tutorial_2021} 
   & LSTM 
   & \textbf{Lorry GPS tracks}\footnotemark (FCD) \space in Belgium with 30s reporting interval,  anonymous IDs (reset daily), timestamp, latitude, longitude, speed, and direction 
   & Trajectories are matched to street network to \textbf{count vehicles per network edge per time step}
   \\  \arrayrulecolor{lightgray}  \cmidrule{1-4}
   Gao et al. (2022) \cite{gao_traffic_2022} 
   & GCN+GRU 
   & Taxi GPS tracks (FCD) in Xi'an, China covering 7.7k taxis for 3 months with a sampling interval of 5–30~s  
   & Trajectories are used to  \textbf{count vehicles per POI (n=100) per time step (hourly)}
   \\  \arrayrulecolor{lightgray}  \cmidrule{1-4}
   Lippert et al. (2022) \cite{lippert_learning_2022} 
   & Recurrent GNN 
   & Bird migration data in the form of simulated trajectories and measurements from the European weather radar network
   & Trajectories are converted \textbf{temporal graphs} with to flows (edges) between Voronoi tessellation cells of radar locations (nodes). 
   \\  \arrayrulecolor{lightgray}  \cmidrule{1-4}
   Xue et al. (2022) \cite{xue_leveraging_2022}
   & Transformers
   & SafeGraph daily POI visit counts in NYC, Dallas and Miami
   & Historical visitation data are translated into natural language sentences to fine-tune pre-trained NLP models 
   \\  \arrayrulecolor{lightgray}  \cmidrule{1-4}  
   Zhang et al. (2020) \cite{zhang_curb-gan_2020} 
   & GAN 
   & Taxi GPS tracks (FCD) in Shenzhen, China for 6 months
   & Trajectories are converted to \textbf{traffic movies} (resolution: 40×50, hourly) with average traffic speed and taxi inflow
   \\ 

   \arrayrulecolor{black}  \bottomrule
 \end{tabularx}
\end{table*}

\end{landscape}

\addtocounter{footnote}{-7}
    \footnotetext{\url{https://coast.noaa.gov/htdata/CMSP/AISDataHandler/2017/index.html }}
\addtocounter{footnote}{1}
    \footnotetext{\url{https://zenodo.org/record/4498410}}
\addtocounter{footnote}{1}
    \footnotetext{\url{https://dma.dk/safety-at-sea/navigationalinformation/ais-data}}
\addtocounter{footnote}{1}
    \footnotetext{\url{http://www.shl-dataset.org/activity-recognition-challenge-2021/}}
\addtocounter{footnote}{1}
    \footnotetext{\url{https://sites.google.com/site/yangdingqi/home/foursquare-dataset?pli=1}}
\addtocounter{footnote}{1}
    \footnotetext{\url{http://www.start.umd.edu/gtd/}}
\addtocounter{footnote}{1}
    \footnotetext{\url{https://github.com/bigdata-ufsc/petry-2020-marc/tree/master/data/foursquare_nyc}}
\addtocounter{footnote}{1}
    \footnotetext{\url{https://www.kaggle.com/datasets/giobbu/belgium-obu}}


\section{Why Deep Learning?}
\label{sec:motivation}

This section attempts to summarize the reasons stated by authors to motivate the use of deep learning (DL) instead of classical machine learning (ML) and discusses the baselines, metrics, and implementations used to measure and compare the performance of DL methods. Finally, this section ends with a critical discussion of the findings.

\subsection{DL Motivations}

Novelty is a crucial factor determining what gets published and what does not. Consequently, we encounter a lot of motivation statements stressing the novelty of applying DL to certain trajectory-related tasks, along the lines of, for example, ``applying CNN for classiﬁcation of trajectory data is relatively unexplored''~\cite{yang_classification_2018} or ``predicting the arrival times with the help of deep learning models has been done in some recent works [...] However, most of these works aim to predict the arrival time for road vehicles [...], and the railroad industry lagged behind''~\cite{yin_deep_2021}.

Beyond the novelty factor, a key motivation for using DL rather than traditional ML is that DL does not rely on hand-crafted features.  Feature engineering for mobility prediction and classification is challenging due to: 
``1) the complex sequential transition regularities exhibited with time-dependent and high-order nature; 
2) the multi-level periodicity of human mobility; and
3) the heterogeneity and sparsity of the collected trajectory data''~\cite{feng_deepmove_2018}. 
Hand-crafted features may therefore run into difficulties capturing the full complex picture necessary for accurate predictions or classifications~\cite{chen_ship_2020,lyu_plug-memory_2022}.

DL's ability to take advantage of large volumes of data to learn complex non-linear spatio-temporal relationships is therefore a key motivating factor. This is particularly stressed by, for example, Derrow-Pinion et al.~\cite{derrow-pinion_eta_2021}, who argue that arrival time prediction ``requires accounting for complex spatio-temporal interactions (modelling both the topological properties of the road network and anticipating events—such as rush hours—that may occur in the future).'' 
On a similar note, Liatsikou et al.~\cite{liatsikou_denoising_2021} argue that DL models for anomaly detection ``can effectively address the challenges of feature extraction, high dimensionality and non-linearity. Moreover, there is no need to explicitly describe a normal pattern for a trajectory and to define the type of the anomaly. [...] the model is trained to reconstruct most of the trajectories included in a dataset, while it fails on the more irregular ones, which are classified as anomalies''
And Buroni et al.~\cite{buroni_tutorial_2021} state that DL models  for road traffic forecasting ``allowed to take advantage of the enormous volume of mobility data to capture the complex non-linear space-temporal relationships governing road traffic.''  Moreover, 
DL may help to accurately ``perform multi-horizon road traffic predictions on large scale transportation networks'' and still ``comply to real-time requirements'' \cite{lv_traffic_2014} (which may not be achievable using established ITS or traffic simulation approaches). 

Indeed, the black-box nature of DL may be advantageous in some use cases dealing with the issue of location privacy. For example, Rao et al.~\cite{rao_lstm-trajgan_2020} argue that current approaches for privacy preservation ``rely heavily on manually designed procedures'' and that ``once the procedure is disclosed, one may have the chance to recover the original trajectory data (e.g., using reverse engineering).'' Furthermore, for synthetic data generation, ``the trade-off between the effectiveness of trajectory privacy protection and the utility for spatial and temporal analyses is still hard to control, and this issue has not been fully discussed or evaluated. Besides, current studies mainly focus on the spatial dimension of trajectory data whereas other semantics (e.g., temporal and thematic attributes) are rarely considered. In fact, these characteristics have been proven to be crucial for trajectory user identiﬁcation''~\cite{rao_lstm-trajgan_2020}.

The before-mentioned advantages notwithstanding, DL may not always be the best approach~\cite{grinsztajn_why_2022}. Reasons include the black-box nature of many DL models~\cite{mokbel2023towards} as well as the large amounts of training data \replaced{and computational}{, computational, and energy} resources needed to train DL models~\cite{shi_thinking_2023}. 
\added{For example, Jonietz et al.~\cite{jonietz2022urban} point out that the "lack of explainability can result in erosion of trust from users who may question the results", which in turn may limit the practical impact of DL developments. The explainability of ML and DL  models remains an open challenge~\cite{mokbel2023towards}. 
While classical ML models often rely on handcrafted features engineered by domain experts, DL models automatically learn features from raw data. In the absence of large, diverse, and representative datasets, DL models may not always capture domain-specific nuances and may be prone to overfitting.
Additionally, “the rise of compute-intensive AI research can have significant, negative environmental impacts"~\cite{lacoste2019quantifying}, depending on how and where the energy for training DL models is generated, stored, and delivered~\cite{lacoste2019quantifying,shi_thinking_2023}. }
The choice of DL models over traditional ML models should therefore be founded in significantly better results than simpler classic ML baselines.

\subsection{Baselines \& Metrics}

In order to evaluate the performance of new methods, it is essential for authors to report benchmark datasets and models as well as the considered metrics for evaluating their experiments. Since the machine learning tasks vary according to the considered mobility use case, comparability can at best be achieved within one use case. 
Therefore, Tables~\ref{table:comparisons}-\ref{table:comparisons2} summarize the different traditional ML and DL methods that have been used as baselines  as well as the metrics  for each  use case. 

The number of ML and DL baselines provided per publication varies widely, as shown in Tables~\ref{table:comparisons}-\ref{table:comparisons2}. While publications on \textit{Traffic volume / crowd flow prediction} and \textit{Next location / destination prediction}  tend to provide numerous baselines for comparison, the number of baselines for \textit{Location classification}, \textit{Trajectory prediction}, and \textit{Anomaly Detection} are much lower. This may hint at a general lack of suitable reference implementations.

Within most use cases, authors tend to adopt similar metrics: 
Classification tasks, such as \textit{Location classification},  \textit{(Sub)trajectory classification} and -- to a degree -- \textit{Next location / destination prediction} strongly rely on F1 scores, accuracy and precision metrics. \added{In addition to providing correct destination predictions, another aspect is to provide the prediction as soon as possible~\cite{gulisano_debs_2018}}.
On the other hand, regression tasks, such \textit{Arrival time prediction}, \textit{Traffic volume / crowd flow prediction} and \textit{Trajectory prediction} tend to rely on RMSE, MAE, and MAPE.
Papers on \textit{Synthetic data generation}, however, do not feature prominent common metrics.

As Tables~\ref{table:comparisons}-\ref{table:comparisons2} show, some trajectory data-specific DL methods, such as VaNext, DeepMove, and LSTM-TrajGAN, have been used repeatedly as baselines to evaluate new methods. 
However, not all reviewed publications provide the source code of the published models to facilitate reuse of the implementation. 
Table~\ref{table:repos} lists the publicly available implementations grouped by the used machine learning library (PyTorch, TensorFlow, or Keras) and ordered by their popularity. It is worth noting that most repositories are one-shot projects, that is, the implementations are not actively worked on and maintained after publication of the paper which likely negatively affects their re-usability.

In contrast, there is a growing number of trajectory analysis libraries which offer essential functionality for machine learning from trajectory data, such as \textit{Trackintel}\footnote{\url{https://github.com/mie-lab/trackintel}}~\cite{martin_trackintel_2023} (e.g., used by~\cite{hong_how_2022}), \textit{MovingPandas}\footnote{\url{http://movingpandas.org}}~\cite{graser_movingpandas_2019} (e.g., used by~\cite{mehri_contextual_2021}) and \textit{scikit-mobility}\footnote{\url{https://scikit-mobility.github.io/scikit-mobility/}}~\cite{pappalardo_scikit-mobility_2021} (e.g., used by~\cite{simini_deep_2021}) as well as spatiotemporal deep learning frameworks, such as TorchGeo\footnote{\url{https://torchgeo.readthedocs.io/en/stable/}}~\cite{stewart_torchgeo_2022} (which focuses on raster data) and 
GeoTorchAI\footnote{\url{https://github.com/wherobots/GeoTorchAI}}~\cite{chowdhury_geotorch_2022} (which includes a bicycle flow prediction example using ST-ResNet~\cite{zhang_deep_2017}).



\begin{landscape}

\begin{table*}[ht]
 \centering
 \scriptsize
 \caption{ML and DL baselines used to compare proposed new methods, including papers that do not provide baseline comparisons \textit{(n/a)}}
 \label{table:comparisons}
   \begin{tabularx}{1.55\textwidth}{X X X X} 
   \toprule
    \textbf{Ref} & \textbf{ML baselines} & \textbf{DL baselines} & \textbf{Metrics} \\ 
\rowcolor{traj}\multicolumn{4}{l}{\color{black}\textbf{Trajectory prediction}} \\ 
  Carroll et al. (2022)~\cite{carroll_towards_2022}
   & Agent-based models
   & \textit{(n/a)}
   & Model Validation Loss
   \\  \arrayrulecolor{lightgray}   \cmidrule{1-4}
 Fan et al. (2022)~\cite{fan_online_2022}
   & ARIMA, prophet
   & \textit{(n/a)}
   & RMSE, MAE, MAPE, Cross Entropy
   \\  \cmidrule{1-4}
  Mehri et al. (2021) \cite{mehri_contextual_2021}  
   & \textit{(n/a)}
   & ordinary LSTM
   & RMSE, point-wise horizontal error
   \\  \cmidrule{1-4}
  Musleh et al. (2022) \cite{musleh_towards_2022} 
   & \textit{(n/a)}
   & \textit{(n/a)}
   & Mean and median shortest Euclidean distance
   \\  \cmidrule{1-4}
  Tritsarolis et al. (2021)~\cite{tritsarolis_online_2021}
   & \textit{(n/a)}
   & \textit{(n/a)}
   & For the co-movement prediction: spatial similarity, temporal similarity, Jaccard similarity 
   \\  \cmidrule{1-4}
  \added{Capobianco et al. (2023)~\cite{capobianco_recurrent_2023}}
   & \added{\textit{(n/a)}}
   & \added{Previous version of their labeled architecture}
   & \added{Average prediction error (APE), average displacement error (ADE)}
   \\
\rowcolor{arrival}\multicolumn{4}{l}{\color{white}\textbf{Arrival time prediction}} 
\\
Derrow-Pinion et al. (2021)~\cite{derrow-pinion_eta_2021}
   & Non-parametric models using \newline real-time \& historical speeds 
   & DeepSets 
   & RMSE
   \\ \arrayrulecolor{lightgray}  \cmidrule{1-4}
Buijse et al. (2021)~\cite{yin_deep_2021}
   & \textit{(n/a)} 
   & \textit{(n/a)}
   &  RMSE, MAE
   \\ \cmidrule{1-4}
Wang et al. (2018)~\cite{wang_when_2018}
   & Average speed, nearest neighbour, GBDT, Perceptron
   & RNN
   & RSME, MAE, MAPE
   \\  
\rowcolor{subtraj}\multicolumn{4}{l}{\color{black}\textbf{(Sub)trajectory classification}} \\  
 Chen et al. (2020) \cite{chen_ship_2020} 
   & KNN, SVM, DT 
   & \textit{(n/a)}
   & F1, Accuracy, AUC 
   \\ \arrayrulecolor{lightgray} \cmidrule{1-4}
 Wang et al. (2021) \cite{wang_locomotion_2021}
   & SHL Challenge summary paper comparing ML \& DL submissions
   & 
   & F1 
   \\   \cmidrule{1-4}
 Yang et al. (2022) \cite{yang_ship_2022} 
   & \textit{(n/a)}
   & \textit{(n/a)}
   & Accuracy, Precision, Recall, F1
   \\ 
\rowcolor{anomaly}\multicolumn{4}{l}{\color{white}\textbf{Anomaly detection }} \\ 
  Liatsikou et al. (2021) \cite{liatsikou_denoising_2021} 
   & Local Outlier Factor (LOF)  
   & Denoising autoencoder with FFNN
   & Precision, Recall, F1
   \\  \arrayrulecolor{lightgray}  \cmidrule{1-4}
 Nguyen et al. (2022) \cite{nguyen_geotracknet_2022}
   & \textit{(n/a)}
   & TREAD, LSTM, VRNN
   & Qualitative manual expert evaluation
   \\  \cmidrule{1-4}
 Singh et al. (2022) \cite{singh_leveraging_2022} 
   & Similarity model 
   & RNN
   & Accuracy, false detection rate
   \\ 
\rowcolor{nextloc}\multicolumn{4}{l}{\color{white}\textbf{Next location / destination prediction }} \\ 
  Feng et al. (2018) \cite{feng_deepmove_2018}
   & HMM
   & RNN
   & Accuracy 
   \\  \arrayrulecolor{lightgray}   \cmidrule{1-4} 
  Gao et al. (2019)~\cite{gao_predicting_2019}
   & HMM 
   & PRME, ST-RNN, Bi-LSTM, POI2Vec, DeepMove
   & RMSE
   \\ \cmidrule{1-4}
  Hong et al. (2022) \cite{hong_how_2022}
   & HMM
   & LSTM, DeepMove, MobTcast, LSTM attn
   & Accuracy, Precision, Recall, F1
   \\ \cmidrule{1-4}
  Li et al. (2020)~\cite{li_predicting_2020}
   & \textit{(n/a)}
   & ST-RNN, MCARNN, DeepMove, VANext
   & Negative Log-Likelihood, Accuracy, Average Percentage Rank
   \\  \cmidrule{1-4}
  Liao et al. (2018) \cite{liao_predicting_2018}
   & Most Frequent, Context- \newline aware hybrid model
   & ST-RNN, Semantic-Enriched-RM, Multi-Task GRU, SCARNN
   & Accuracy, Negative Log-Likelihood, Mean Average Precision
   \\  \cmidrule{1-4}
  Liu et al. (2016) \cite{liu_predicting_2016}
   & \textit{(n/a)}
   & RNN, ST-RNN
   & F1, accuracy, precision, recall, AUC
   \\ \cmidrule{1-4}
  Tenzer et al. (2022) \cite{tenzer_meta-learning_2022}
   & \textit{(n/a)}
   & RNN, Attention-based BiLSTM
   & Mean Haversine Distance (MHD)
   \\ \cmidrule{1-4}
  \added{Gulisano et al. (2018) \cite{gulisano_debs_2018}}
   & \added{DEBS 2018 Grand Challenge paper}
   & 
   & \added{Average time span between a correct prediction and the arrival at the destination}
   \\ 
  \arrayrulecolor{black} \bottomrule
 \end{tabularx}
\end{table*}

\begin{table*}[ht]
 \centering
 \scriptsize
 \caption{ML and DL baselines used to compare proposed new methods, including papers that do not provide baseline comparisons \textit{(n/a)} -- continued}
 \label{table:comparisons2}
   \begin{tabularx}{1.55\textwidth}{X X X X} 
   \toprule
    \textbf{Ref} & \textbf{ML baselines} & \textbf{DL baselines} & \textbf{Metrics} \\ 

\rowcolor{synth}\multicolumn{4}{l}{\color{black}\textbf{Synthetic data generation }} \\ 
 Rao et al. (2020) \cite{rao_lstm-trajgan_2020}   
   & \textit{(n/a)}
   & LSTM-TrajGAN 
   & Accuracy, Precision, Recall, F1, Hausdorff Distance, Jaccard Index, temporal visit probability distribution 
   \\ \arrayrulecolor{lightgray}  \cmidrule{1-4}
  Simini et al. (2021) \cite{simini_deep_2021} 
   & Gravity Model (G), Nonlinear Gravity Model (NG)
   & Multi-Feature Gravity Model (MFG)
   & Common Part of Commuters (CPC), Relative improvement w.r.t. G
   \\ \cmidrule{1-4}
  Zhang et al. (2022) \cite{zhang_factorized_2022} 
   & Rule-based models
   & LSTM, IGMM-GAN, VAE
   & Mean Distance Error (MDE),  Maximum Mean Discrepancy (MMD), novelty score, Violation Score (VS)
   \\

\rowcolor{location}\multicolumn{4}{l}{\color{white}\textbf{Location classification}} 
   \\ 
 Altan et al. (2022)~\cite{altan_discovering_2022}
   & \textit{(n/a)}
   & \textit{(n/a)}
   & Precision, recall, F-score 
   \\ \arrayrulecolor{lightgray} \cmidrule{1-4}
 Lyu et al. (2022)~\cite{lyu_plug-memory_2022}
   & \textit{(n/a)}
   & \textit{(n/a)}
   & F1-scores (micro and macro)
   \\ \cmidrule{1-4}
 Yang et al. (2018)~\cite{yang_classification_2018} 
   & \textit{(n/a)}
   & \textit{(n/a)}
   & Precision
   \\ 

\rowcolor{volume}\multicolumn{4}{l}{\color{black}\textbf{Traffic volume / crowd flow prediction}}   
\\ 
 Buroni et al. (2021)~\cite{buroni_tutorial_2021} 
   & Seasonal Window Persistence Model (SW)
   & LSTM (directed, iterated and with encoder-decoder approach)
   & RMSE, MAE
   \\ \arrayrulecolor{lightgray}  \cmidrule{1-4} 
 Gao et al. (2022)~\cite{gao_traffic_2022}
   & ARIMA, SVR 
   & GRU, DCRNN, TGNN
   & RMSE, MAE, Accuracy
   \\ \cmidrule{1-4} 
 Li et al. (2021) \cite{li_prediction_2021} 
   & ARIMA, KNN, GBDT
   & LSTM, AIP-PM, AIP-PGM, ASTGCN
   & RMSE, MAE
   \\ \cmidrule{1-4}
  Lippert et al. (2022)~\cite{lippert_learning_2022}
   & Historical Average (HA), \newline GAM, GBT
   & \textit{(n/a)}
   & RMSE
   \\ \cmidrule{1-4}
  Wang et al. (2022) \cite{wang_periodic_2022} 
   & Historical Average (HA)
   & DeepST, ST-ResNet, ConvLSTM, DeepSTN, Graph WaveNet, DeepLGR
   & RMSE, MAE
   \\ \cmidrule{1-4}
  Zhang et al. (2020)~\cite{zhang_curb-gan_2020} 
   & Spatial smoothing
   & ConvLSTM, FC-SA, CNN-SA, FC-LSTM, CNN-LSTM, DyConv-LSTM
   &  RMSE, MAPE
   \\  \cmidrule{1-4}
  Xue et al. (2022) \cite{xue_leveraging_2022}
   & Linear Regression
   & GRU, GRUAtt, Transformer
   & RMSE, MAE
   \\ 

   \\

   \arrayrulecolor{black} \bottomrule
 \end{tabularx}
\end{table*}

\end{landscape}

\begin{table*}[ht]
 \centering
 \scriptsize
 \caption{Neural network implementations published together with their reviewed papers, ordered by ML library and stars. The stars column lists the number of Github stars, Kaggle upvotes, or Figshare/Zenodo downloads of the respective repository as of July 2023.}
 \label{table:repos}
   \begin{tabularx}{1\textwidth}{ >{\hsize=2pt}X >{\hsize=20pt}X  >{\hsize=75pt}X  >{\hsize=175pt}X  >{\hsize=25pt}X X } 
      \toprule
    & \textbf{Ref} & \textbf{Model/System Name} & \textbf{Code Repository} & \textbf{Stars} & \textbf{Last} \newline \textbf{updated}  \\ 
   \cmidrule{1-6}
\multicolumn{6}{l}{\textbf{PyTorch}} \\
  \rowcolor{nextloc!10}\cellcolor{nextloc} & 
  \cite{feng_deepmove_2018} &  DeepMove & \url{https://github.com/vonfeng/DeepMove} & 131 & 08/19
  \\  
  \rowcolor{arrival!10} \cellcolor{arrival} & \cite{wang_when_2018} &  DeepTTE with \newline GEO-Conv layer & \url{https://github.com/UrbComp/DeepTTE} & 127 & 06/20 
  \\ 
  \rowcolor{synth!10} \cellcolor{synth} & \cite{simini_deep_2021} & DeepGravity & \url{https://github.com/scikit-mobility/DeepGravity} & 57 & 12/21 
  \\ 
  \rowcolor{volume!10}\cellcolor{volume} &  \cite{zhang_curb-gan_2020} & Curb-GAN & \url{https://github.com/Curb-GAN/Curb-GAN} & 25 & 02/20 
  \\ 
  \rowcolor{nextloc!10}\cellcolor{nextloc} &  \cite{hong_how_2022} & \textit{n/a} & \url{https://github.com/mie-lab/location-mode-prediction} & 9 & 07/23
  \\ 
  \rowcolor{volume!10}\cellcolor{volume} &  \cite{lippert_learning_2022} & FluxRGNN & \url{https://github.com/FionaLippert/FluxRGNN} \newline \url{https://zenodo.org/record/6921595} & 3 & 07/22
  \\ 
  \rowcolor{arrival!10} \cellcolor{arrival} & \cite{yin_deep_2021} & Deep Train Arrival \newline Time Estimator & \url{https://github.com/basbuijse/train-arrival-time-estimator} & 2 & 07/21
  \\ 
  \rowcolor{traj!10} \cellcolor{traj} & 
  \cite{fan_online_2022} & \textit{n/a} & \url{https://github.com/fanzipei/crowd-context-prediction/tree/master} & 0 & 02/21
  \\ 
  \rowcolor{volume!10}\cellcolor{volume} &  \cite{xue_leveraging_2022} & AuxMobLCast & \url{https://github.com/cruiseresearchgroup/AuxMobLCast} & 0 & 11/22
  \\ 
  \rowcolor{synth!10}\cellcolor{synth} &  \cite{zhang_factorized_2022} & TrajGen & \url{https://github.com/tongjiyiming/TrajGen} & 0 & 10/20
  \\ 
\multicolumn{4}{l}{\textbf{TensorFlow}} \\
  \rowcolor{volume!10}\cellcolor{volume} &  \cite{li_prediction_2021} & \textit{n/a} & \url{https://figshare.com/articles/dataset/Prediction_of_human_activity_intensity_using_the_interactions_in_physical_and_social_spaces_through_graph_convolutional_networks/11829306/1} & 122 & 03/21
  \\ 
  \rowcolor{anomaly!10}\cellcolor{anomaly} &  \cite{nguyen_geotracknet_2022} & GeoTrackNet & \url{https://github.com/CIA-Oceanix/GeoTrackNet} & 67 & 10/20
  \\ 
  \rowcolor{anomaly!10} \cellcolor{anomaly} & \cite{singh_leveraging_2022} & Uncertainty EDL \newline Graph & \url{https://github.com/sansastra/uncertainty_edl_graph} & 8 & 07/21
  \\  
\multicolumn{4}{l}{\textbf{Keras}} \\
  \rowcolor{synth!10}\cellcolor{synth} &  \cite{rao_lstm-trajgan_2020} & LSTM-TrajGAN & \url{https://github.com/GeoDS/LSTM-TrajGAN} & 42 & 01/23
  \\ 
  \rowcolor{subtraj!10}\cellcolor{subtraj} &  \cite{chen_ship_2020} & CNN-SMMC & \url{https://github.com/rechardchen123/Ship_movement_classification_from_AIS} & 19 & 04/20
  \\ 
  \rowcolor{volume!10} \cellcolor{volume} & \cite{buroni_tutorial_2021} & Tutorial on traffic forecasting with DL & \url{https://www.kaggle.com/code/giobbu/lstm-encoder-decoder-tensorflow} & 9 & 09/21
  \\ 
  \rowcolor{anomaly!10} \cellcolor{anomaly} & \cite{liatsikou_denoising_2021} & BMDA anomaly detection & \url{https://github.com/marialiatsikou/BMDA_anomaly_detection} & 0 & 02/21
   \\
   \arrayrulecolor{black}   \bottomrule  
 \end{tabularx}
\end{table*}

\subsection{Discussion}

The lack of common benchmarks methods and datasets makes it difficult to compare methods and to systematically advance the research field. For example, Wang et al.~\cite{wang_locomotion_2021}, remark that ``to date, most research groups assess the performance of their algorithms using their own datasets on their own recognition tasks. These tasks often differ in the sensor modalities or in the allowed recognition latency.'' Additionally, there is a lack of benchmark datasets for most use cases (as noted, e.g., regarding location classification by Altan et al.~\cite{altan_discovering_2022}). 
For some use cases, researchers, therefore, resort to synthetic data. For example, Liatsikou et al. (2021) create their own synthetic trajectory anomalies to evaluate their model \cite{liatsikou_denoising_2021}. 

Another issue is the lack of agreed evaluation approaches and metrics. 
While papers introducing new DL methods -- quite naturally -- claim superior performance over existing ML/DL methods, data challenges, such as the SHL Challenge on traffic flow prediction, paint a less clear picture, with regular ML models often outperforming DL models on key metrics, such as F1 score, train time, and test time. For example, Wang et al.~\cite{wang_locomotion_2021} compare 7 ML SHL Challenge submissions (including XGBoost, RF, DT and ensembles of classifiers) and 8 DL submissions (including CNN, LSTM, CNN+RNN, Transformer, and AdaNet), coming to the conclusion that ``ML outperforms DL in global. ML has a similar higher bound as DL, while achieving a much higher lower bound. The smaller dynamic range verifies the better robustness of ML. While DL achieves the highest F1 score (75.4\%), it is only 1.1 percentage points higher than the best ML approach (74.3\%).'' Furthermore, as would be expected ``DL takes much more time for training than ML, and also takes more time for testing.''~\cite{wang_locomotion_2021}
Regarding DL's independence from hand-crafted features, it is particularly interesting that ``Feature-based DL approaches achieve a much higher F1 score (75.4\%) than raw-data-based approaches (43.4\%).''~\cite{wang_locomotion_2021} 

Since the majority of papers are not directly reproducible (either due to the lack of data, code, or both) or prohibitively expensive to reproduce, verification of stated claims remains a challenge. 
Typical pitfalls faced when evaluating models for mobility use cases include~\cite{widhalm_top_2018}: 1) different datasets employing often (undocumented) pre-processing pipelines 2) unnoticed over-fitting due to spatial and or temporal auto-correlation and 3) the accuracy/generality trade-off due to idealized conditions and lack of variation in the data.  In particular, Widhalm et al. (2018)~\cite{widhalm_top_2018} show that evaluation with a random training/test split suggests a considerably higher accuracy than with proper backtesting (reducing the F1 score from 96\% to 84\% when accounting for auto-correlation and down to 54\% when transferring the model to different conditions). Therefore, they argue that results reported in publications may not represent reliable indicators for the future performance in real-world applications, where generality and robustness are essential.

\section{Trends}
\label{sec:trends}

While previous sections provided a detailed review of use cases and types of deep neural networks for trajectory data, their focus on recent work is insufficient to comprehensively understand longer-term trends in this field. 
Therefore, in this section, we investigate these longer-term trends by through the following research questions: How has the DL research on mobility trajectory application domains evolved in recent years? Which DL architectures are being utilized in research, and how have they changed over time?

To answer these questions, we performed a systematic literature review (SLR) following the PRISMA method~\cite{page2021prisma} for data collection and selection. PRISMA is a widely used guideline for conducting SLRs. The search was carried out on 'Google Scholar' and 'IEEE Xplore' using the following keywords:

\begin{itemize}
\footnotesize
    \item Google Scholar: \texttt{"Deep Learning" AND "Trajectories" AND "Mobility" -"survey"}
    \item Google Scholar: \texttt{"Machine Learning" AND "Trajectories" AND "Mobility"}
    \item IEEE Explor: \texttt{"Deep Learning" AND "Trajectories" AND "Mobility"}
\end{itemize}

and limited to the timeline between 2018 and 2023. The total number of initially retrieved publications was 337. The following steps were performed to select relevant papers: First, we discarded duplicate articles (8 publications) within and across databases. After that, we reviewed titles and abstracts as well as other parts of the publication if necessary to manually classify each publication regarding its use case and machine learning technology. Publications that either do not address one of the previously introduced eight use cases as listed in Tab.~\ref{table:overview} (39 publications) or do not apply neural networks are discarded (24 publications). In the next step, all publications not published in a journal, at a conference, as a book chapter, or as a preprint (14 publications) were excluded. We also considered preprints, as many machine learning publications are initially released as preprints before being presented in a journal or at a conference much later. We hope that this will allow us to assess current trends more accurately.
In addition, review papers were excluded from further analysis (13 publications). Finally, we included the papers analyzed in the previous sections (36 publications), which yielded a total of 275 relevant papers. The relevant papers comprise 111 articles, 114 journals, 25 preprints, and 0 book chapters.

\begin{figure}[h!]
  \centering
  \includegraphics[width=1\linewidth]{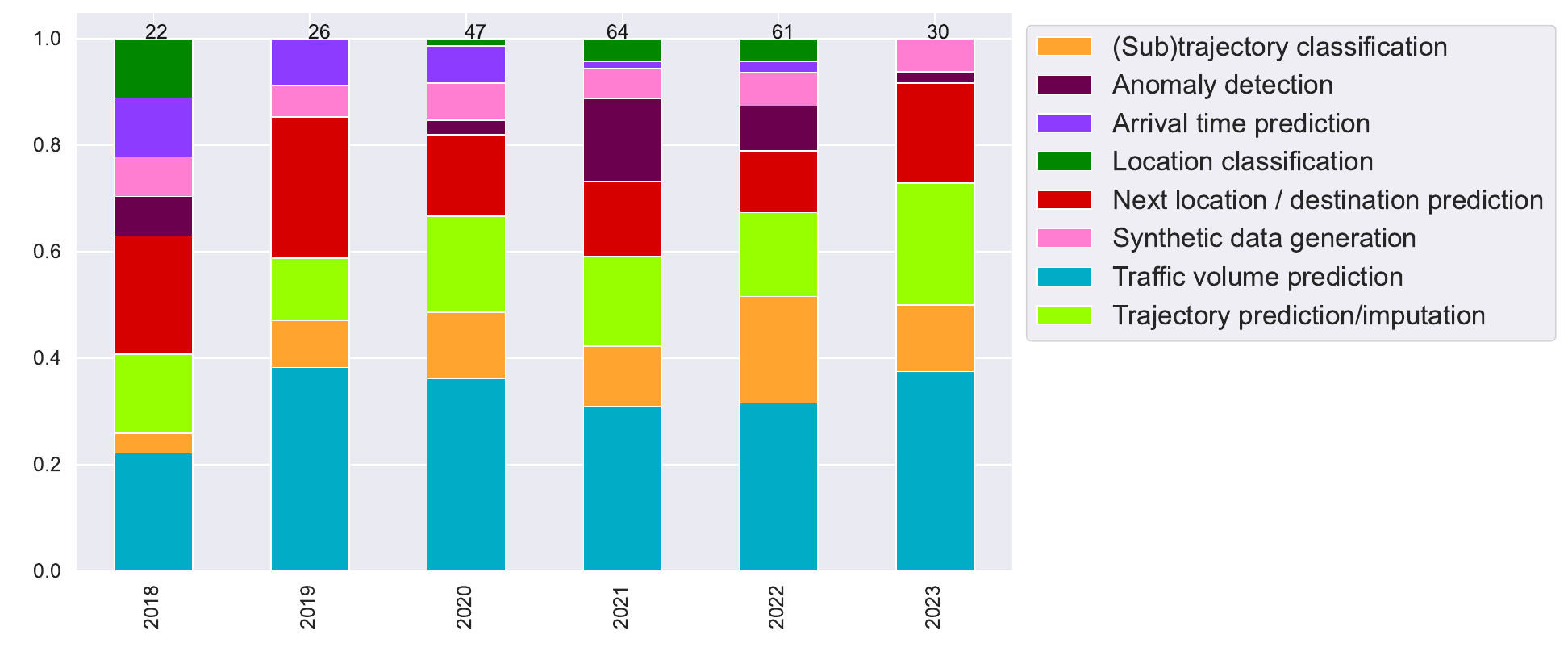}
  \caption{Portions of publications by use case from 2018 to 2023. Total number of publications per year is displayed at the top of each bar.}
  \label{fig:UseCaseTrendBar}
\end{figure}

Figure~\ref{fig:UseCaseTrendBar} shows the trend analysis results for the eight previously introduced use cases. 
By far, the most popular use case is \textit{'Traffic volume~/~crowd flow prediction'}, responsible for approximately ~33\% of publications. The next most common use cases are \textit{'Trajectory prediction/ imputation'}, followed by \textit{'Next location~/~destination prediction'} and \textit{'(Sub)trajectory classification'}. On the other hand, the least common use cases are \textit{'Anomaly detection'}, \textit{'Synthetic data generation'}, \textit{'Arrival time prediction'}, and \textit{'Location classification'}.  It is worth noting that the numbers for 2023 only represent publications up until the time of writing (June 2023) and that publications can address more than one use case. Please refer to the Section Appendix for the absolute number of publications per use case.

To effectively analyze the vast array of diverse neural network architectures extracted from the relevant publications, we categorized and clustered them to be able to discern prevailing trends and patterns. We applied the following categorizations:

\begin{enumerate}
\footnotesize
    \item RNN -- Vanilla RNN, LSTM, or GRU
    \item GNN -- Graphs and graph learning methods
    \item CNN -- Convolutional networks
    \item AE  -- Autoencoders (Dense, LSTM, or CNNs)
    \item Attention  -- Attention mechanisms, including transformers
    \item GAN -- Generative adversarial methods
    \item DRL -- Deep reinforcement learning
    \item CRNN -- Hybrid architectures combining CNN and RNN methods (e.g., LSTM or GRU)
    \item FNN -- Fully connected dense networks
    \item Other DL -- Any other deep learning architecture, such as Capsule Net, Gravity Net, etc.
\end{enumerate}

This categorization is more fine-grained than the summary in Table~\ref{table:overview} to allow for better differentiation between the less common NNs and their respective trends. 

\begin{figure}[h!]
  \centering
  \includegraphics[width=0.8\linewidth]{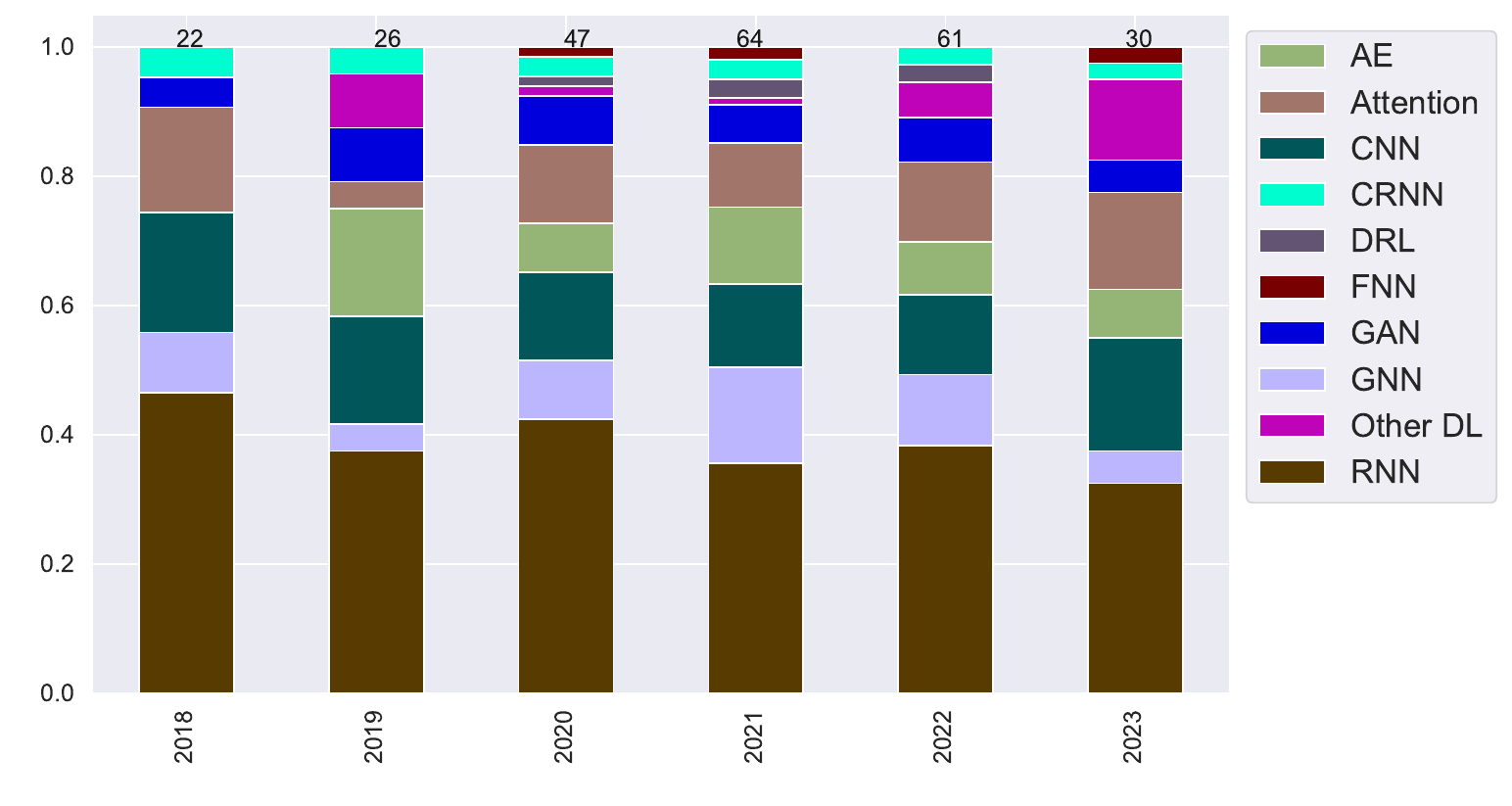}
  \caption{Portions of publications by NN design from 2018 to 2023. Total number of publications per year is displayed at the top of each bar.}
  \label{fig:DLtrend_bar}
\end{figure}

Even more than the use case trends, this neural network design trends analysis summarized in Figure~\ref{fig:DLtrend_bar} is dominated by a single class: RNNs, accounting for approximately 39\% of publications.  
The second most popular option for the early years were CNNs, which decreased in popularity in 2020. CNNs have since been overtaken by attention mechanisms and -- potentially -- GNNs (but the 2023 data here is still inconclusive). Autoencoders (AE) comprised approximately 18\% of the publications in 2019 and spiked again in 2021, which seems to be decreasing in popularity since. FNNs, CRNNs, DRL, GAN, and other DL methods have remained niche applications. Note that publications can integrate multiple neural network designs.

\begin{figure}[h!]
  \centering
  \includegraphics[width=1\linewidth]{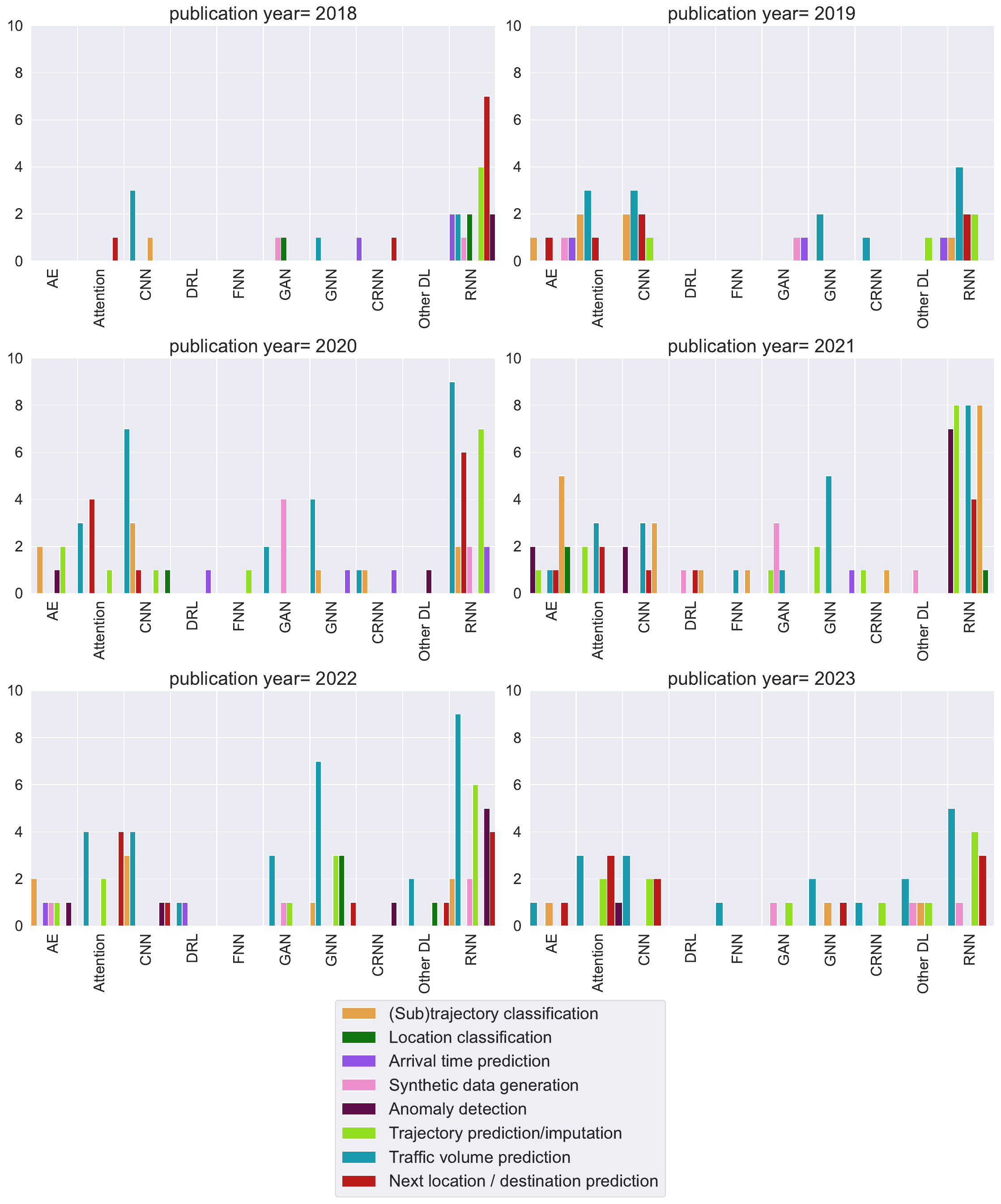}
  \caption{Overview of the deep learning approaches per use cases  from 2016 to 2023}
  \label{fig:grid}
\end{figure}

Bringing both neural network design and use case perspective together, we also analyzed the techniques employed for each use case over the years, as shown in Figure~\ref{fig:grid}. This shows the prevalent use of RNNs for tasks such as traffic volume prediction, trajectory prediction, and next location/final destination prediction.
Furthermore, we observed an emerging trend where GNNs are increasingly being adopted as a viable solution for traffic volume and trajectory prediction use cases. Similarly, CNNs have gained popularity as a practical approach for traffic volume prediction. However, in trajectory classification tasks, they serve as the second choice after RNNs.

\section{Conclusion}
\label{sec:conclusion}

In this work, we reviewed deep learning-based research focusing on trajectory data in mobility use cases. In most cases, even if dense raw trajectory data is used in the process, it is not ingested directly for training the neural networks. Instead, data engineering steps are applied that convert trajectories into more compact representations of individual trajectories (sparse trajectories) or aggregations of multiple trajectories. This aggregated trajectory data is commonly presented as time series of vectors,  graphs, or images (movies). Since these data engineering steps are a recurring need in mobility data science, we expect further uptake of trajectory analysis libraries, such as \textit{Trackintel}, \textit{MovingPandas}, and \textit{scikit-mobility} since these libraries implement many common trajectory generalization, aggregation, and analysis methods and their focus on scientific software engineering aims at reusability and long(er)-term availability of the software libraries. This is an essential step towards more sustainable mobility data science, since most of the implementations summarized in Table~\ref{table:repos} have not been substantially updated/maintained since being published which reduces their reusability since the implementations will -- sooner or later -- become incompatible with newer versions of the underlying data science libraries.


Future research should address the issues of model transferability, benchmark availability, and model explainability. 
Current work rarely addresses the issue of model transferability. Since most existing global ML models ``cannot perform well locally, or be transferred to study similar problems in other regions''\cite{janowicz_six_2022}, transferability should be considered when evaluating or comparing models.
Additionally, developed models, even for the same application and trajectory type, are difficult to evaluate (for example, due to the lack of ground truth for anomaly detection) and to compare due to different datasets and applied metrics. Therefore, more open benchmark datasets are needed. 
Finally, to better understand the inner workings of neural networks and to ensure the trustworthiness of model decisions, explainability should play a more crucial role in model development.

\section*{Declarations} 

\subsection*{Competing interests}  
We, the authors, declare that we have no competing interests as defined by Springer, or other interests that might be perceived to influence the results and/or discussion reported in this paper.
  
\subsection*{Authors' contributions} 
All authors contributed to the study’s conception and design. All authors contributed to the in-depth qualitative review of recent work. A.G. wrote the original draft and compiled the tables and Figure~1. A.J and A.W. performed the quantitative literature review and prepared Figures~2-6. All authors participated in contributing to the text and the content of the manuscript, including revisions and edits. All authors approve of the content of the manuscript. All authors reviewed the manuscript.
  
\subsection*{Funding} 
This work is mainly funded by the  EU's Horizon Europe research and innovation program under Grant No. 101070279 MobiSpaces and No. 101093051 EMERALDS, and the EU's Horizon 2020  program under Grant No. 101021797 STARLIGHT.
  
\subsection*{Availability of data and materials}  
The dataset and code to reproduce the quantitative literature review are available at \url{https://github.com/anahid1988/MobilityDL}.

%
%
%
 \bibliographystyle{splncs04}
 \bibliography{paper}

\section*{Appendix}

\subsection*{Absolute numbers of literature review results}

\begin{figure}[h!]
  \centering
  \includegraphics[width=1\linewidth]{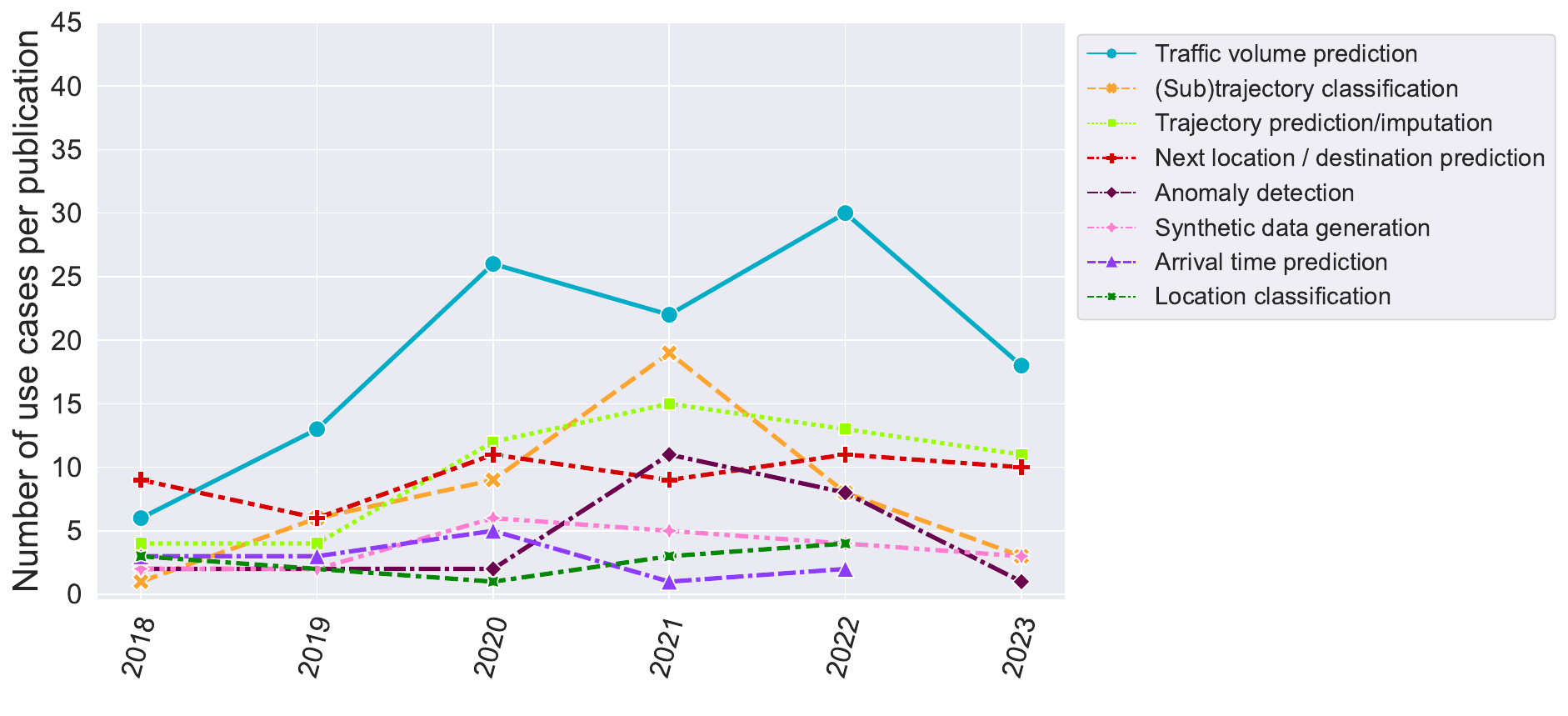}
  \caption{Publication counts by use case from 2018 to 2023.}
  \label{fig:UseCaseTrend}
\end{figure}

\begin{figure}[h!]
  \centering
  \includegraphics[width=0.8\linewidth]{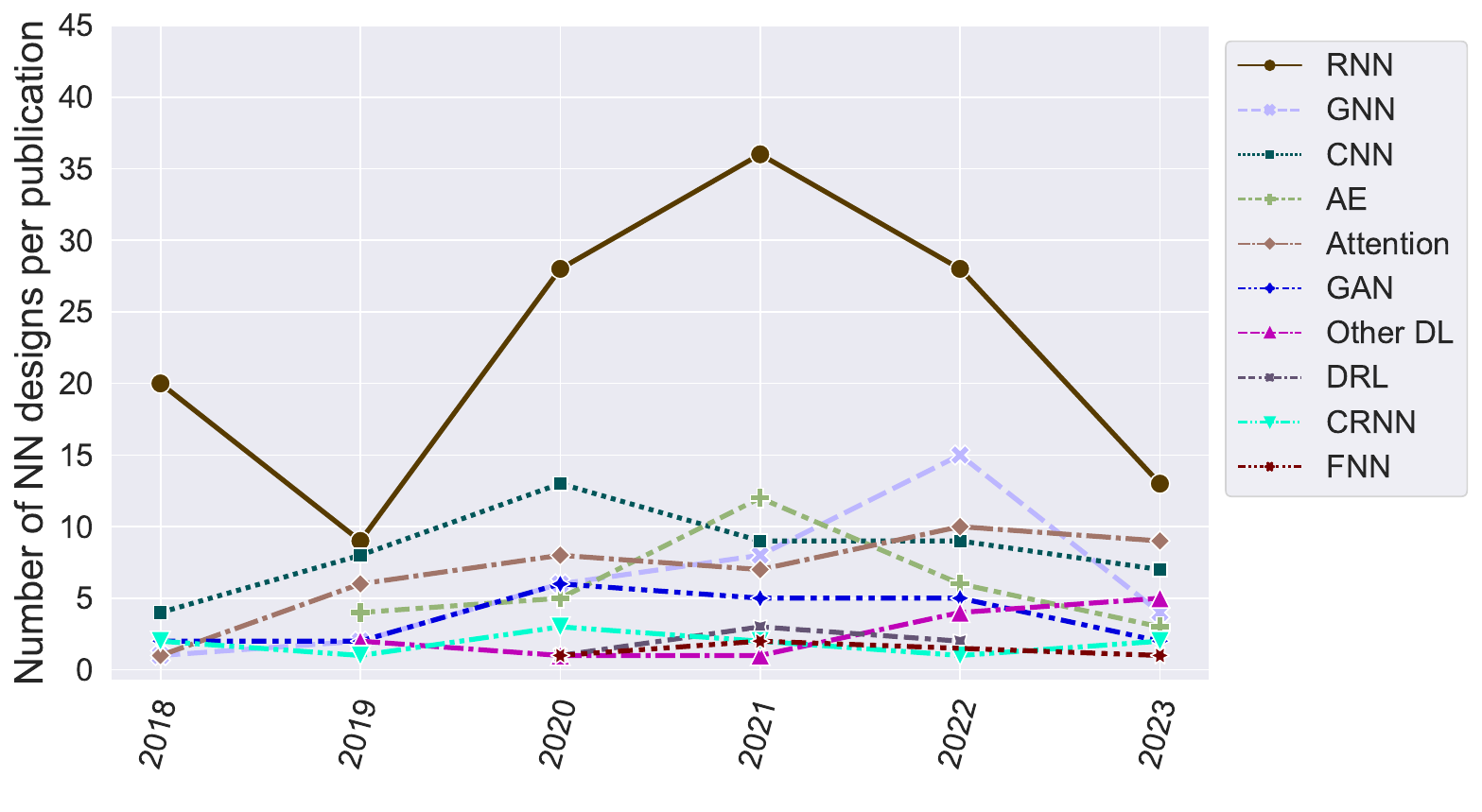}
  \caption{Publication counts by NN design from 2018 to 2023.}
  \label{fig:DLtrend}
\end{figure}

\subsection*{Glossary}
\begin{itemize}
\footnotesize
    \item AE -- Autoencoder
    \item AIS -- Automatic Identification System
    \item \added{APE -- Average prediction error}
    \item \added{ADE -- Average displacement error}
    \item \added{AUC -- Area under the ROC curve} 
    \item CDR -- Call Detail Records
    \item CNN -- Convolutional Neural Network
    \item COG -- Course over ground
    \item DNN -- Deep Neural Network
    \item DL -- Deep Learning
    \item DT -- Decision Tree 
    \item FNN –- Fully connected dense networks
    \item FFNN -- Feed-Forward Neural Network
    \item FCD -- Floating Car Data
    \item GAN -- Generative Adversarial Network
    \item GBDT -- Gradient-boosted Decision Tree 
    \item GeoAI -- Geospatial Artificial Intelligence
    \item GIS -- Geographic Information Science
    \item GNN -- Graph Neural Network
    \item GPS -- Global Positioning System, often used synonymously for all GNSS (incl. Galileo, GLONASS, and Beidou) 
    \item HMM -- Hidden Markov Models
    \item KNN -- K-Nearest Neighbours
    \item LSTM -- Long Short-Term Memory
    \item \added{MAE -- Mean absolute error }
    \item \added{MAPE -- Mean absolute percentage error}
    \item ML -- Machine Learning
    \item MLP -- Multilayer Perceptron
    \item NLP -- Natural Language Processing
    \item OBU -- On-board Unit
    \item OD --  Origin-Destination
    \item POI -- Point of Interest
    \item PRME -- Personalized Ranking Metric Embedding
    \item RF -- Random Forest 
    \item RNN -- Recurrent Neural Network
    \item SAE -- Stacked Autoencoder
    \item SAN -- Self-Attention Network
    \item SOG -- Speed over ground
    \item SVM -- Support Vector Machine
    \item VAE -- Variational AutoEncoder 
    \item XGBoost -- Extreme Gradient Boost 
\end{itemize}

\end{document}